\documentclass{article}

\usepackage{microtype}
\usepackage{graphicx}
\usepackage{subcaption}
\usepackage{booktabs} 

\usepackage{hyperref}

\usepackage[preprint]{icml2026}

\usepackage{amsmath}
\usepackage{amssymb}
\usepackage{mathtools}
\usepackage{amsthm}

\usepackage[capitalize,noabbrev]{cleveref}

\theoremstyle{plain}

\theoremstyle{definition}

\theoremstyle{remark}

\newcommand{\benchname}{DynaHOI-10M}
\newcommand{\suitename}{DynaHOI-Gym}
\newcommand{\baselinename}{ObAct}
\newcommand\ie{\textit{i.e.}}

\usepackage[textsize=tiny]{todonotes}

\icmltitlerunning{DynaHOI: Benchmarking Hand-Object Interaction for Dynamic Target}

\usepackage{url}
\usepackage{graphicx} 
\usepackage{listings}
\usepackage{xcolor}
\usepackage{caption}
\usepackage{pifont}
\usepackage{multirow}
\usepackage{enumitem}

\usepackage{titletoc}

\usepackage[ruled,algo2e]{algorithm2e}
\usepackage{algorithm}
\usepackage{algorithmic}
\usepackage{comment}
\usepackage{nicefrac} 

\usepackage{amssymb}
\usepackage{mathtools}
\usepackage{amsthm}
\usepackage{comment}
\usepackage{minitoc}
\usepackage{graphicx}
\usepackage{booktabs}
\usepackage{color}

\usepackage{tcolorbox}

\usepackage{cleveref}
\usepackage{cuted}
\usepackage{xpatch}
\usepackage{array}
\usepackage{multirow}
\usepackage{graphicx}
\usepackage{xcolor,colortbl}
\usepackage{pifont}
\usepackage{enumitem}
\usepackage{xcolor}
\usepackage[dvipsnames]{xcolor}

\definecolor{Gray}{gray}{0.9}

\usepackage{booktabs,multirow,colortbl,xcolor,pifont}
\newcommand{\yes}{\textcolor{green!60!black}{\ding{51}}} 
\newcommand{\no}{\textcolor{red!70!black}{\ding{55}}}  

\usepackage{fancyvrb}
\usepackage{mdframed}

\begin{document}

\twocolumn[
  \icmltitle{DynaHOI: Benchmarking Hand-Object Interaction for Dynamic Target}



  \icmlsetsymbol{equal}{*}

  \begin{icmlauthorlist}
    \icmlauthor{Bocheng Hu}{zju}
    \icmlauthor{Zhonghan Zhao}{zju,ailab}
    \icmlauthor{Kaiyue Zhou}{sichuan}
    \icmlauthor{Hongwei Wang}{zju}
    \icmlauthor{Gaoang Wang}{zju}
  \end{icmlauthorlist}
  \vskip 0.1in

  \icmlaffiliation{ailab}{Shanghai AI Lab, Shanghai, China}
  \icmlaffiliation{zju}{Zhejiang University, Hangzhou, China}
  \icmlaffiliation{sichuan}{Sichuan Provincial Digital Human Center, Chengdu Minto Technology Co., Ltd., Chengdu, China}

  \icmlcorrespondingauthor{Gaoang Wang}{gaoangwang@intl.zju.edu.cn}

  \icmlkeywords{Machine Learning, ICML}

   {\centering
  \includegraphics[width=0.95\linewidth]{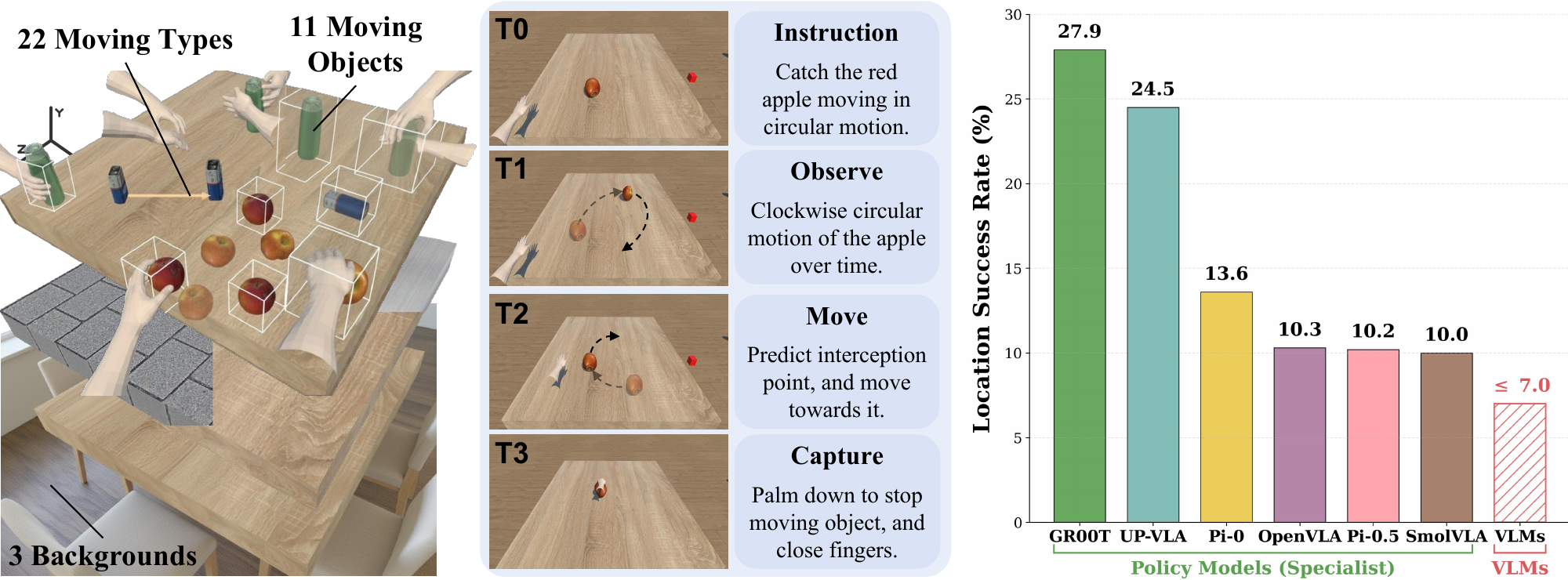}
  \captionof{figure}{Dynamic capture in \benchname: 11 moving objects and 22 moving types \textbf{(left)}. Given an instruction and a short observation window, the model must predict the interception point and capture the moving target \textbf{(middle)}. Existing policy models and generalist VLMs achieve low localization success rate on our benchmark, highlighting the challenge of motion-aware anticipation \textbf{(right)}.}
  \label{fig:teaser}
  \par}
    
   \vskip 0.3in
]



\printAffiliationsAndNotice{}  

\begin{abstract}
  Most existing hand motion generation benchmarks for hand–object interaction (HOI) focus on static objects, leaving dynamic scenarios with moving targets and time-critical coordination largely untested. To address this gap, we introduce the \textbf{\suitename}, a unified online closed-loop platform with  parameterized motion generators and rollout-based metrics for dynamic capture evaluation. Built on \suitename, we release \textbf{\benchname}, a large-scale benchmark with \textbf{10M} frames and \textbf{180K} hand capture trajectories, whose target motions are organized into \textbf{8} major categories and \textbf{22} fine-grained subcategories. We also provide a simple observe-before-act baseline (\baselinename) that integrates short-term observations with the current frame via spatiotemporal attention to predict actions, achieving an \textbf{8.1\% improvement} in location success rate. Our code is available at \url{https://github.com/HuBocheng/DynaHOI}.
\end{abstract}

\section{Introduction}\label{sec:intro}
\begin{table*}[t]
\centering
\small
\caption{\textbf{Comparison of hand--object benchmarks for \emph{dynamic} capture.}
Prior datasets are mostly static and annotation-heavy, typically evaluated by framewise alignment to ground truth.
\textbf{DynaHOI-Gym / DynaHOI-10M} instead provides parameterized target-motion generators and \emph{online} closed-loop evaluation with rollout-based success and trajectory-quality metrics.}
\resizebox{\textwidth}{!}{
\renewcommand{\arraystretch}{1.15}
\begin{tabular}{lccccccc}
\toprule
\multirow{2}{*}{Benchmark} 
& \multicolumn{2}{c}{Protocol} 
& \multicolumn{3}{c}{Data Collection} 
& \multicolumn{2}{c}{Dataset Scale} \\
\cmidrule(lr){2-3}\cmidrule(lr){4-6}\cmidrule(lr){7-8}
& Online eval. & Controlled targets. 
& Automated collect. & Parameterized variability & Human annot. 
& \#Frames & \#Traj \\
\midrule

GRAB~\citep{GRAB_2020}          
& \no & \no 
& \no & \no & \yes 
& 1.0M & 6.5K \\

HOI4D~\citep{liu2022hoi4d}          
& \no & \no 
& \no & \no & \yes 
& 2.4M & 4K \\

AssemblyHands~\citep{AssemblyHands}          
& \no & \no 
& \no & \no & \yes 
& 3.03M & 62 \\

Ego-Exo4D~\citep{egoexo4d2024}          
& \no & \no 
& \no & \no & \yes 
& --- & --- \\

TACO~\citep{liu2024taco}          
& \no & \no 
& \no & \no & \yes 
& 4.7M & 2.3K \\

OakInk2~\citep{oakink2}          
& \no & \no 
& \no & \no & \yes 
& 4.01M & 2.8K \\

HOT3D~\citep{HOT3D}          
& \no & \no 
& \no & \no & \yes 
& 3.7M & 4.1K \\

GigaHand~\citep{fu2025gigahands} 
& \no & \no 
& \no & \no & \yes 
& 2.4M & 13.9K \\
\rowcolor{gray!10}
\textbf{Ours (DynaHOI-Gym / DynaHOI-10M)} 
& \yes & \yes 
& \yes & \yes & \no 
& \textbf{10.0M} & \textbf{180K} \\

\bottomrule
\end{tabular}
}
\label{tab:compare_hoibench}
\end{table*}
Hand-object interaction in the real world is fundamentally perceptual and predictive: humans observe a moving target with their eyes, reason over its motion in the brain, and issue precise motor commands to complete the task. Tasks involving \emph{moving} objects place higher demands on short-horizon prediction and real-time reasoning. Although existing hand-motion generation models have demonstrated strong performance on various benchmarks, nearly all benchmarks focus exclusively on static objects, making them misaligned with both practical requirements and the rapidly advancing landscape of policy models~\citep{GRAB_2020,liu2022hoi4d,egoexo4d2024,oakink2,fu2025gigahands}. This leaves a critical blind spot: can current models accurately forecast where a moving target will be in the near future and carry out the manipulation instructed by a human? As illustrated in \Cref{fig:teaser}, we construct a benchmark in which targets follow well-defined motion patterns. We evaluate both action-outputting policy models (VLA, diffusion policies) and general-purpose VLMs, and, as shown in the right portion of the figure, their performance exhibits a substantial gap from the ground-truth controller.

Directly evaluating dynamic capture in the physical world is difficult: collecting human demonstrations is costly, target motion is hard to reproducibly control, and large-scale data generation and batch testing are impractical. To make dynamic capture measurable and reproducible, we develop the \textbf{\suitename}, a unified, online, closed-loop evaluation platform for dynamic hand-object interaction. It provides parameterized target-motion generators for moving targets and evaluates models from episode rollouts using success rate, trajectory quality, and task completion speed, rather than framewise alignment to mocap ground truth.

On top of \suitename, we construct \textbf{\benchname}, a large-scale benchmark dedicated to dynamic hand-motion generation for moving-target capture. \benchname~contains \textbf{10M frames} across \textbf{180K} capture trajectories, with target motion patterns organized into \textbf{8 major categories} and \textbf{22 fine-grained subcategories}. Beyond motion diversity, the benchmark features a wide range of objects, backgrounds, and textures, supporting \emph{both} visual diversity and physically grounded motion diversity within a unified evaluation setup. Unlike prior work that emphasizes frame-by-frame action prediction, each episode follows an \emph{observe-before-act} rollout: an observation phase where the hand watches the moving target to infer its motion, followed by a control phase where the model predicts an interception point and generates actions to complete the capture.

In brief, our main contributions are summarized as follows:
\begin{enumerate}[leftmargin=*]
\item \textbf{\suitename~platform.} We develop \suitename, a unified suite for dynamic capture that supports trajectory generation and collection for diverse objects under pre-defined motion types. It also provides \emph{closed-loop} online evaluation with metrics covering success rate, trajectory quality, and task completion speed.
\item \textbf{\benchname~benchmark.} We build \benchname, a large-scale benchmark with \textbf{10M frames} across \textbf{180K} capture trajectories, covering \textbf{8} motion categories and \textbf{22} subcategories. It supports both frame-by-frame and observe-before-act rollouts.
\item \textbf{Observe-before-act baseline.} We also provide an observe-before-act baseline (ObAct) that first observes target motion, then predicts actions for interception and capture. Our method delivers consistent improvements over the prior diffusion-based models \textbf{across all seven metrics}, including an \textbf{8.1\%} gain in location success rate.

\end{enumerate}



\begin{figure*}[t]
\centering
    \includegraphics[width=\linewidth]{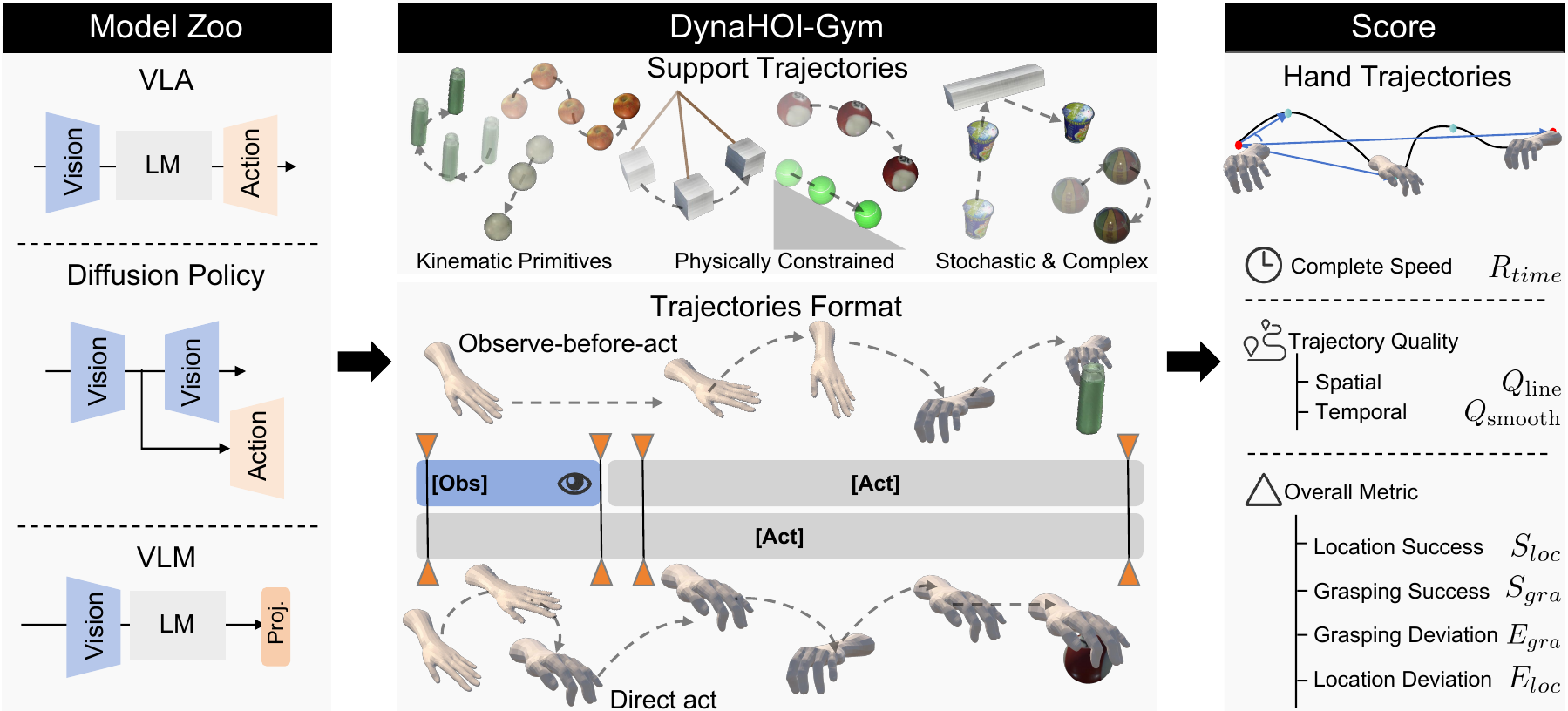}
    \caption{\textbf{The framework of \benchname~benchmark.} 
Model zoo with three families, \textbf{VLA}, \textbf{diffusion policies}, and \textbf{VLM}-based controllers. 
\emph{Middle:} \suitename~supports trajectories (kinematic primitives, physically constrained, stochastic \& complex) and two rollout formats: \emph{observe-before-act} and \emph{direct act}. 
\emph{Right:} Multi-dimensional scoring includes runtime $R_{\text{time}}$, trajectory quality (spatial $Q_{\text{line}}$, temporal $Q_{\text{smooth}}$), and overall metrics (location/grasp success $S_{\text{loc}},S_{\text{gra}}$; deviations $E_{\text{loc}},E_{\text{gra}}$).}
    \label{fig:framework}
\end{figure*}

\section{Related Work}

\noindent\textbf{Hand Action Generation.}
Recent HOI methods increasingly treat interaction as sequence generation rather than per-frame pose recovery, with diffusion and language-conditioned models becoming the dominant thread. InterHandGen~\cite{lee2024interhandgen} models two-hand contact geometry via a cascaded reverse-diffusion prior that factorizes joint distributions to improve plausibility/diversity in close interactions . Building on diffusion backbones for controllable motion, HandDiffuse~\cite{HandDiffuse} formulates generative controllers for two-hand interactions and introduces interaction-aware objectives to reduce artifacts in long motions. For hand–object interaction, Text2HOI~\cite{cha2024text2hoi} pioneers text-guided 3D HOI sequence generation by decomposing the problem into contact-map synthesis + contact-conditioned diffusion motion generation, improving physical plausibility and generalization to unseen objects . Moving toward long-horizon compositionality, HOIGPT~\cite{huang_etal_cvpr25} discretizes HOI sequences with a physically grounded tokenizer (hand/object VQ-VAE) and trains a motion-aware language model to perform bidirectional HOI and text modeling. OpenHOI~\cite{zhang2025openhoi} pushes further to open-world, open-vocabulary HOI synthesis by using a 3D MLLM for affordance grounding and instruction decomposition. 
Only very recent efforts begin to explore policy-level generation for dexterous hands under richer formulations: DexHandDiff investigates diffusion-based controllers for contact-rich hand manipulation \cite{Liang_2025_CVPR}, while Being-M0 extends VLA models to fine-grained whole-body motion generation and Being-H0 further extends them to hand–object interaction~\cite{wang2025scaling,beingbeyond2025beingh0}. Despite these advances, a systematic evaluation of hand action generation under dynamic targets remains unexplored.

\noindent\textbf{Benchmarks for Hand–Object Interaction.} A variety of benchmarks have been introduced to evaluate 3D hand–object interactions. Early datasets provided precise single-frame annotations of hand and object pose in contact, like ContactPose~\cite{brahmbhatt2020contactpose} captured 2.9M RGB-D images of 2,306 grasps with detailed hand–object contact maps. Building on these, larger-scale motion-capture datasets now cover diverse interaction scenarios. DexYCB~\cite{chao:cvpr2021} focuses on human grasps of everyday objects with accurate 3D hand and object trajectories, while TACO~\cite{liu2024taco} benchmarks generalizable bimanual manipulation with diverse tool–action–object triplets. More recent datasets emphasize broader tasks: OakInk~\cite{oakink2} and HOI4D~\cite{liu2022hoi4d} curate egocentric, category-level manipulations with many object instances, ARCTIC~\cite{fan2023arctic}captures large-scale two-handed dexterous manipulation of articulated objects; and GigaHand~\cite{fu2025gigahands} provides a massive, 51-view, multi-camera dataset of bimanual activities.

\section{Benchmark}
Before introducing our task formulation and evaluation framework, we summarize in \Cref{tab:compare_hoibench} how existing hand–object benchmarks differ in evaluation protocol, target controllability, and data collection paradigms. Most prior datasets emphasize offline reconstruction of static interactions, whereas our benchmark is designed around online, closed-loop evaluation under parameterized target motion.

\begin{figure*}[t]
\centering
    \includegraphics[width=1\linewidth]{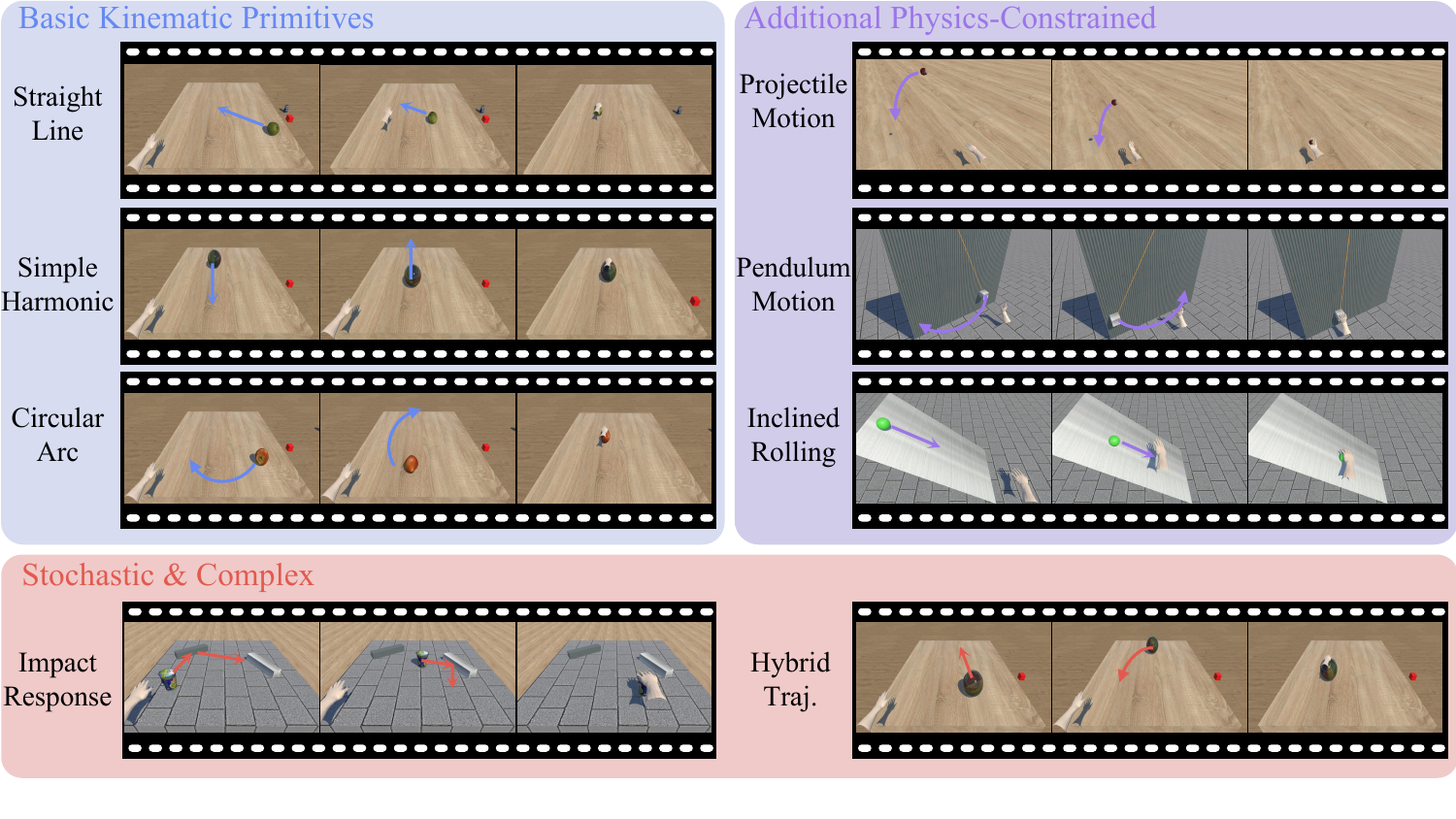}\\
    \caption{\textbf{Motion diversity in the \benchname~benchmark.}
    \benchname~spans 3 categories and 8 major motion types, ranging from kinematic primitives to physics-constrained and stochastic dynamics.
    The illustrated trajectories demonstrate rich and diverse manipulation behaviors over moving targets, enabling comprehensive evaluation of dynamic object grasping.}
    \label{fig:demo}
\end{figure*}

\subsection{Scene and Task Design}
\label{sec:task_design}
We study \textbf{Dynamic Capture}, where a model controls an 18-DoF five-fingered hand to \emph{intercept} and \emph{capture} a moving target object in 3D. Each episode provides a natural-language instruction and an egocentric visual stream. The target follows a pre-defined continuous motion trajectory, while the hand must coordinate palm translation and finger articulation to achieve timely capture.

At each time step $t$, the policy receives an observation
\begin{equation}
o_t = \{ I_t,\; h_t,\; q_t,\; k_t,\; x^{\text{text}} \},
\end{equation}
where $I_t \in \mathbb{R}^{H \times W \times 3}$ is the egocentric RGB image, $h_t \in \mathbb{R}^3$ is the palm  position, $q_t \in \mathbb{R}^{15}$ are the 15 finger joint angles, $k_t$ denotes the 6D fingertip keypoint poses, and $x^{\text{text}}$ is the task description.

The policy outputs an 18-dimensional control command
\begin{equation}
a_t = (a^{\text{loc}}_t,\; a^{\text{gras}}_t), \quad
a^{\text{loc}}_t \in \mathbb{R}^{3},\; a^{\text{gras}}_t \in \mathbb{R}^{15},
\end{equation}
where $a^{\text{loc}}_t$ controls the palm-center translation for interception and $a^{\text{gras}}_t$ controls finger joint rotations for capture.




\subsection{\suitename}
To facilitate online validation of generated motions, rather than limiting evaluation to offline residuals against ground truth, we develop the \suitename~using the Unity physics engine. It functions as a closed-loop simulation platform that allows agents to interact directly with the environment, ensuring that evaluation metrics better reflect real-world applicability.

\paragraph{Dynamic Grasp Trajectory Generation.}
\suitename~automates the creation of diverse interaction scenarios through three core modules:
\begin{enumerate}
    \item \textbf{Parametric Motion Synthesis:} Generates trajectories for diverse object categories by varying motion primitives and physical parameters.
    \item \textbf{Adaptive Grasping Logic:} Generates interception trajectories with capture actions across diverse objects and target motion states, ensuring accurate localization and stable capture under dynamic conditions.
    \item \textbf{Scalable Data Pipeline:} An automated framework that enables the high-throughput collection of large-scale interaction data with synchronized annotations.
\end{enumerate}

\paragraph{Unified Evaluation Interface.}  
As shown in \Cref{fig:framework}, \suitename~functions as a middleware connecting the model with the simulator. It standardizes the data flow by providing the model with real-time visual feedback and proprioceptive states (hand pose) while parsing the model's 18-DoF output into low-level simulator control signals. Crucially, the interface enforces strict temporal synchronization, ensuring that the simulator's physics engine and the model's inference clock remain aligned during online evaluation.

\paragraph{Observe-Before-Act Mechanism.}  
Unlike static scenarios, dynamic capture emphasizes temporal prediction and anticipation. To this end, \suitename~introduces an ``observe-before-act" protocol in both data collection and evaluation. As shown in the middle panel of \Cref{fig:framework}, the hand remains fixed while the object moves according to predefined dynamics. This observation window allows the model to perceive temporal variations and learn motion patterns. 
\suitename~supports both \textbf{observe-before-act} and \textbf{direct-act} formats, accommodating evaluation for both static and dynamic tasks.

\paragraph{Hierarchical Evaluation Metrics.}  
In dynamic capture task, single-dimensional evaluation cannot capture overall model performance. As illustrated in the right panel of \Cref{fig:framework}, we design a hierarchical metric system with three levels: overall success, trajectory quality, and completion speed. Formula definitions are provided in \Cref{sec:exp}.

\begin{figure*}[t]
\centering
    \includegraphics[width=1\linewidth]{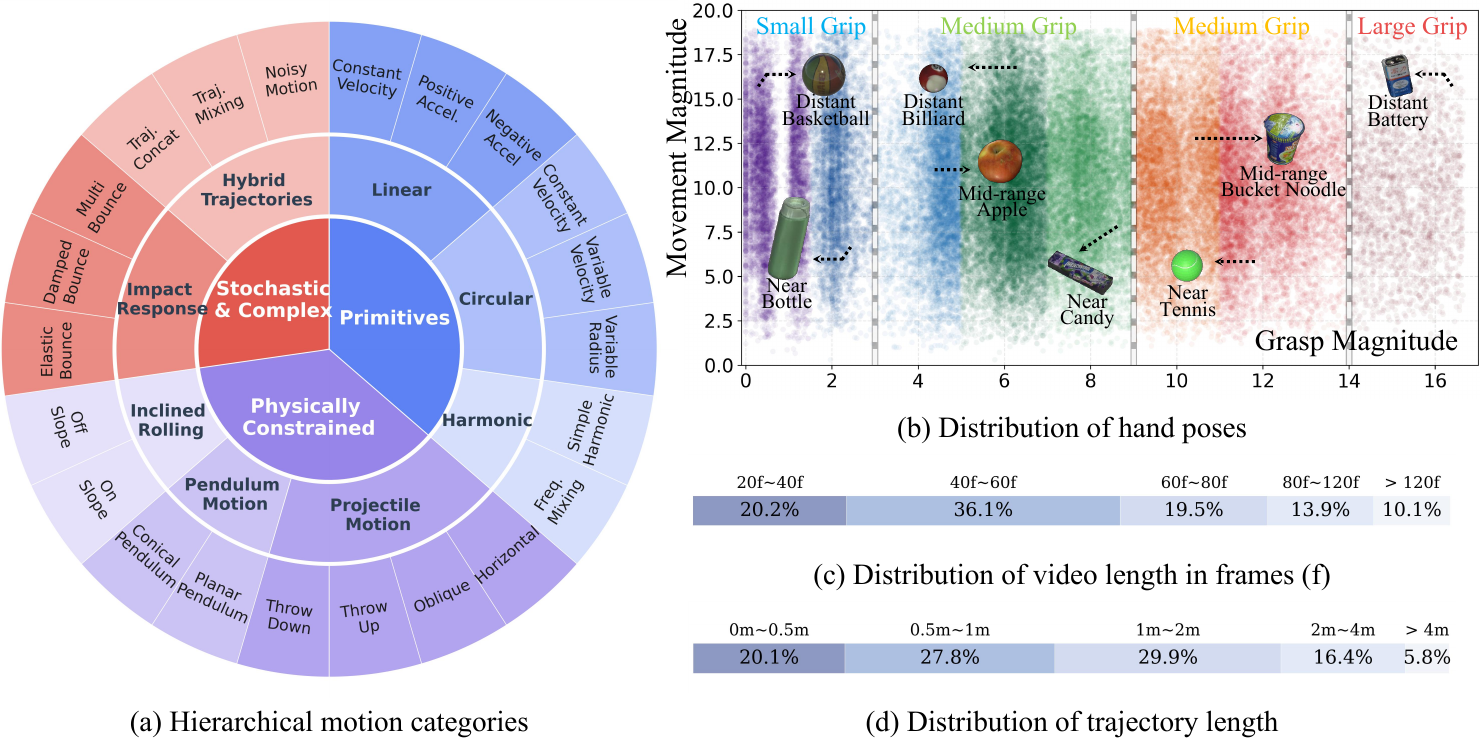}
    \caption{\textbf{Data statistics and diversity of \benchname.}
    (a) Object motions are organized into a \textbf{3}-level hierarchy with \textbf{8} major categories and \textbf{22} fine-grained subcategories.
    (b) \textbf{Hand poses cover diverse object scales}: larger objects correspond to smaller grasp magnitudes and smaller objects to larger grasp magnitudes; each object supports both near- and far-range manipulation.
    (c--d) Distributions of episode durations (frames) and trajectory lengths across the benchmark.}
    \label{fig:stats1}
\end{figure*}

\subsection{\benchname~Benchmark}

\paragraph{Dataset Statistics.}
As shown in \Cref{fig:demo}, \benchname~organizes target dynamics into a three-level hierarchy: \textit{Basic Kinematic Primitives}, \textit{Additional Physics-Constrained} motions, and \textit{Stochastic \& Complex} dynamics. This design covers 8 major motion types (Straight Line, Simple Harmonic, Circular Arc, Projectile Motion, Pendulum Motion, Inclined Rolling, Impact Response, and Hybrid Trajectory), ranging from simple kinematic patterns to physics-governed and stochastic behaviors, ensuring systematic motion diversity for dynamic capture evaluation.

Beyond the major types, \Cref{fig:stats1}(a) further decomposes them into 22 fine-grained subcategories, each equipped with a compositional parameter space. For instance, circular motions vary by center position, radius, and angular velocity, while projectile motions vary by peak height, initial speed, and launch angle. Combined with 11 object categories and diverse scene backgrounds and textures, \benchname~results in a visually and physically diverse benchmark with 180K episodes and 10M image frames, spanning approximately 140 hours.

\paragraph{Diversity Analysis.}
\Cref{fig:stats1}(b) reports the joint distribution of grasp magnitude and movement magnitude, covering a broad spectrum from precision grips to large-aperture grasps under varying motion ranges, reflecting diverse interaction modes between the hand and moving targets. As shown in~\Cref{fig:stats1}(c), Episode durations concentrate around 40--60 frames (36.1\%) while retaining substantial coverage of shorter (20--40 frames, 20.2\%) and longer horizons (60--80 frames, 19.5\%; 80--120 frames, 13.9\%; $>$120 frames, 10.1\%), yielding evaluations that jointly stress motion inference (observation sufficiency) and sustained timing-aware control. Notably, longer videos do not necessarily imply longer motion paths. As shown in~\Cref{fig:stats1}(d), traveled distances are dominated by moderate ranges (0.5--2m, 47.7\%) but include near-field (0--0.5m, 20.1\%) and long-range motions (2--4m, 16.4\%; $>$4m, 5.8\%), covering both rapid local interception and long-horizon pursuit regimes.

\paragraph{Model Zoo.}
We construct a comprehensive Model Zoo on \benchname~featuring 12 representative models, ranging from specialized policy models to generalist vision-language models (VLMs). Specifically, we evaluate six policy models, comprising two autoregressive models (OpenVLA~\citep{openvla2024}, UP-VLA~\citep{upvla2025}) and four diffusion-based models (GR00T-N1.5~\citep{gr00tn1_2025}, SmolVLA~\citep{shukor2025smolvla}, Pi-0~\citep{black2024pi0}, Pi-0.5~\citep{physicalintelligence2025pi05}), covering mainstream architectures. Furthermore, to assess the potential of general-purpose multimodal intelligence, we evaluate six mainstream VLMs, including four proprietary models (Gemini-3 Pro, GPT-5.1, Qwen3-Max, Grok4.1) and two open-weights models (Qwen3-VL 235B~\cite{Qwen3-VL}, InternVL-3.5 241B~\cite{InternVL3.5}). 
\begin{figure}[t]
\centering
    \includegraphics[width=\linewidth]{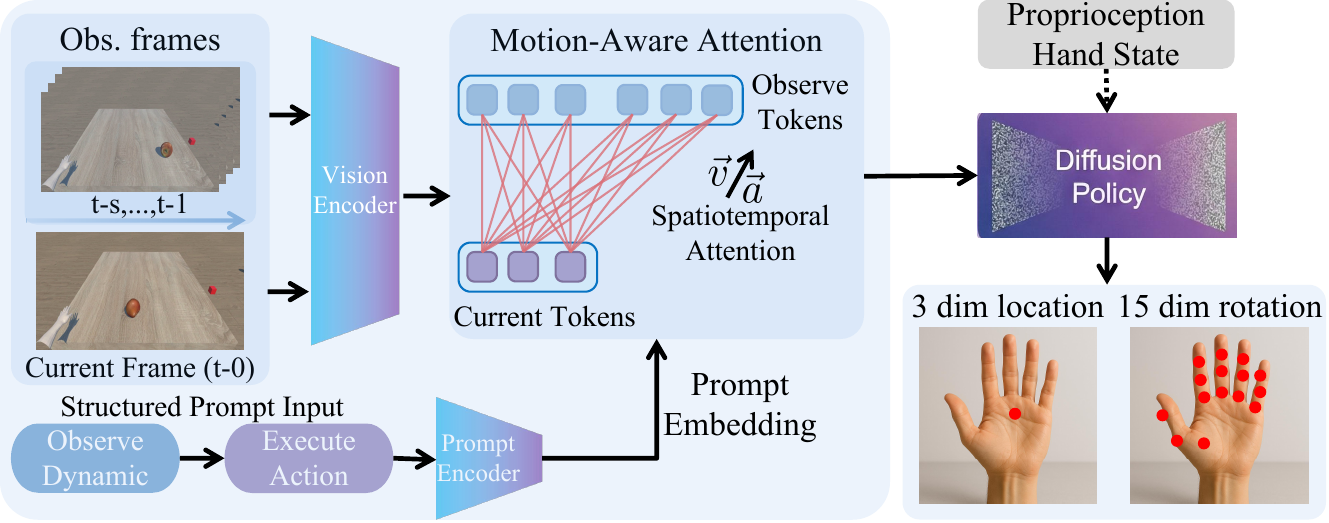}
    \caption{ObAct incorporates observation frames and spatiotemporal attention to condition action prediction on object dynamics.}
    \label{fig:baseline_framework}
    \vspace{-6pt}
\end{figure}
\subsection{Observe-Before-Act Baseline}
Standard policy models typically predict actions from a single visual frame, which limits their ability to reason about object motion in dynamic grasping scenarios. Since instantaneous observations lack temporal cues, such models are inherently unable to infer target velocity or acceleration, often resulting in reactive and poorly timed behaviors.

To address this limitation, we propose an observe-before-act baseline, \textbf{ObAct}, which conditions action generation on a short temporal observation window. As shown in \Cref{fig:baseline_framework}, ObAct uses the backbone VLM to encode egocentric images and task text into visual–language tokens, then applies \emph{spatiotemporal attention} over historical observation ($s$ frames) and the current frame to implicitly capture target motion dynamics. The resulting motion-aware representation is fused with the current proprioceptive state and injected as conditioning into a diffusion policy that reuses GR00T-N1.5’s action head and weights to output continuous hand actions.

\section{Experiments}
\subsection{Experiment Setup}
All experiments are conducted on the proposed \suitename, which is implemented on the Unity engine (editor version 2022.3.58f1c1). For reproducibility, the internal clock is disabled and the simulator advances with fixed-step intervals to ensure consistent timing. 
Model training and evaluation are deployed on a distributed GPU cluster with CUDA 12.4 and 4 $\times$ A100 GPUs (80GB).

For a fair comparison, we reproduce all policy models using official implementations and minimally fine-tune action-related components on \benchname~training set to align with our hand-control space.
For VLMs, we run closed-loop online rollouts by streaming the current RGB observation and 18-D hand state to the model. 
Similar to VLABench~\cite{zhang2024vlabench}, we prompt the VLM to output a parameterized skill program (a sequence of skills with associated parameters) along with an explicit low-level rollout of the next $T$ steps as 18-D control vectors in a constrained JSON format. \suitename~parses and executes them to compute task outcomes and trajectory-level metrics. 
All evaluations use a fixed random seed to ensure reproducibility (Testing details are in Appendix~\ref{app:TestDetail}).

\subsection{Evaluation metrics}
\label{sec:evaluation_metrics}

\paragraph{Overall Metrics.}
This level directly reflects whether the task is completed as well as the overall error.

We report four overall metrics: \textbf{localization success rate} ($S_{loc}$), the fraction of trajectories in which the hand comes within a distance threshold of the target; \textbf{grasping success rate} ($S_{gra}$), the fraction of trajectories that achieve a successful grasp \emph{conditioned on successful localization}; \textbf{localization error} ($E_{loc}$), the minimum distance between the palm center and the target object over the execution; and \textbf{grasping error} ($E_{gra}$), the minimum distance between the fingertips and the target surface over the execution.

\paragraph{Trajectory Quality Metrics.}
We evaluate temporal smoothness and spatial linearity.

For \textbf{temporal smoothness}, given a trajectory $\{p_t\}_{t=1}^{N}\subset\mathbb{R}^3$, we define step lengths
$d_t=\|p_{t+1}-p_t\|_2$ and $\mathbf d=(d_1,\ldots,d_{N-1})$.
We measure step consistency by the coefficient of variation ($\mathcal{CV}$):
\[
Q_{\text{smooth}} = \frac{1}{1 + \mathcal{CV}(d)}\in [0,1], \quad
CV(d)=
\begin{cases}
\frac{\sigma_d}{\mu_d}, & \mu_d\neq 0,\\
0, & \mu_d=0.
\end{cases}
\]
Here, $\mu_d$ and $\sigma_d$ denote the mean and standard deviation of $\{d_t\}$. Smaller variation yields higher $Q_{\text{smooth}}$, indicating smoother trajectories.


For \textbf{spatial linearity}, let segment directions be
$\hat{s}_t=(p_{t+1}-p_t)/\|p_{t+1}-p_t\|_2$.
When $\|p_N-p_1\|_2>0$, the overall direction is
$\hat{v}=(p_N-p_1)/\|p_N-p_1\|_2$ and we compute
\[
Q_{\text{line}}=\frac{1}{N-1}\sum_{t=1}^{N-1}\hat{s}_t\cdot \hat{v}\in[-1,1].
\]
If $\|p_N-p_1\|_2=0$, we set $Q_{\text{line}}=0$.
Higher $Q_{\text{line}}$ indicates a straighter trajectory, whereas lower values suggest unnecessary curvature or deviation from the overall direction.

\paragraph{Completion Speed Metric.}
Let $N$ be the total frames in the test video and $T$ is the frame index at task completion. The time score (higher scores indicate faster completion) is:
\[
R_{\text{time}} = 1 - \frac{T}{N}, \quad R_{\text{time}} \in [0,1].
\]

\subsection{Evaluation on Model Zoo}
\label{sec:exp}
\paragraph{Evaluation on Policy Models.}

\begin{table*}[t]
\centering
\small
\caption{\textbf{Online evaluation of mainstream policy models.}
$S_{\text{loc}}/S_{\text{gra}}$ are success rates, $E_{\text{loc}}/E_{\text{gra}}$ are minimum overall/endpoint distance.
$Q_{\text{smooth}}$ and $Q_{\text{line}}$ measure trajectory smoothness and line-adherence.
$R_{\text{time}}$ is normalized completion speed. $\uparrow$ and $\downarrow$ indicate higher/lower values are better. 
\textbf{Bold} / \underline{underline} denote best / second-best among learned policies; the \textit{GT} row is an oracle upper bound.}
\renewcommand{\arraystretch}{1.10}
\resizebox{\linewidth}{!}{%
\begin{tabular}{l l c | cc | cc | cc | c}
\toprule
\multirow{2}{*}{Model} & \multirow{2}{*}{Architecture} & \multirow{2}{*}{Params} &
\multicolumn{2}{c|}{Localization} &
\multicolumn{2}{c|}{Grasping} &
\multicolumn{2}{c|}{Trajectory Quality} &
\multicolumn{1}{c}{Runtime} \\
\cmidrule(lr){4-5}\cmidrule(lr){6-7}\cmidrule(lr){8-9}\cmidrule(lr){10-10}
& & & $S_{\text{loc}}$ (\%) $\uparrow$ & $E_{\text{loc}}$ $\downarrow$
& $S_{\text{gra}}$ (\%) $\uparrow$ & $E_{\text{gra}}$ $\downarrow$
& $Q_{\text{smooth}}$ $\uparrow$ & $Q_{\text{line}}$ $\uparrow$
& $R_{\text{time}}$ $\uparrow$ \\
\midrule
\rowcolor{gray!08}
GT  & ---             & ---  & 100.00 & 0.16 & 100.00 & 0.09 & 0.90 & 0.96 & 0.75 \\
\midrule
OpenVLA & \multirow{2}{*}{AutoRegressive} & 7B   & 10.30  & 3.23 & 0.80 & 2.83 & 0.41 & \underline{0.46} & 0.04 \\
UP-VLA   &                            & 2B & 24.50 & \underline{0.84} & 0.40 & 0.75 & \textbf{0.46} & \textbf{0.47} & \underline{0.13} \\
\midrule
GR00T-N1.5   & \multirow{4}{*}{Diffusion} & 3B & \underline{27.90} & 0.91 & \underline{3.50} & \underline{0.68} & 0.27 & 0.12 & 0.09 \\
SmolVLA &                                  & 500M & 10.00  & 2.14 & 0.50 & 1.72 & 0.41 & 0.20 & 0.05 \\
Pi-0 &                                  & 4B & 13.60  & 2.35 & 0.60 & 2.03 & 0.27 & -0.05 & 0.03 \\
Pi-0.5 &                                  & 4B & 10.20  & 2.91 & 0.10 & 2.47 & 0.32 & 0.18 & 0.04 \\
\midrule
ObAct (Ours) &            Diffusion                      & 3B & \textbf{36.00}  & \textbf{0.65} & \textbf{4.60} & \textbf{0.41} & \underline{0.43} & 0.22 & \textbf{0.20} \\
\bottomrule
\end{tabular}
}
\label{tab:exp:vla}
\end{table*}
As shown in \Cref{tab:exp:vla}, dynamic capture remains far from solved. Even the best model (GR00T-N1.5) attains only $27.90\%$ localization success rate ($S_{loc}$), indicating that reliably aligning with a moving target is still a primary bottleneck. More strikingly, once localization succeeds, grasping is still rarely successful: the best conditional grasping success rate is merely $S_{gra}=3.50\%$ (GR00T-N1.5), suggesting that contact-rich closing under motion remains highly challenging.

Trajectory quality further reveals a clear trade-off. Diffusion policies attain higher task success but exhibit unstable paths (e.g., GR00T-N1.5: $Q_{smooth}=0.27$, $Q_{line}=0.12$; Pi-0/Pi-0.5 show near-zero/negative $Q_{line}$), whereas the autoregressive UP-VLA produces the cleanest geometry and trajectories ($E_{loc}=0.84$, $Q_{smooth}=0.46$, $Q_{line}=0.47$) yet with limited success ($S_{loc}=24.50\%$, $S_{gra}=0.40\%$).

Finally, our ObAct baseline consistently outperforms all prior diffusion-based policy models across \emph{all seven} metrics, while achieving trajectory quality comparable to autoregressive-based models (e.g., near-second-best $Q_{\text{smooth}}$). Notably, a lightweight temporal mechanism yields substantial gain in motion-aware interception: $S_{\text{loc}}$ improves by $8.1\%$ over the strongest diffusion baseline. The highest $R_{\text{time}}$ further suggests that temporal context accelerates target locking and localization. Since ObAct \emph{reuses} GR00T-N1.5’s action head and weights, the improvements can be attributed primarily to better temporal modeling rather than better low-level control.

\paragraph{Evaluation on VLM Models.}

\begin{table}[t]
\centering
\large
\caption{\textbf{Online evaluation of mainstream VLMs.} Metrics follow \Cref{tab:exp:vla}; $S_{\text{loc}}^{*}$ uses a more lenient localization threshold (See Appendix~\ref{app:TestDetail} for details).}
\renewcommand{\arraystretch}{1.10}
\resizebox{\linewidth}{!}{%
\begin{tabular}{l | cc | cc | cc | c}
\toprule
\multirow{2}{*}{Model} &
\multicolumn{2}{c|}{Localization} &
\multicolumn{2}{c|}{Grasping} &
\multicolumn{2}{c|}{Trajectory Quality} &
\multicolumn{1}{c}{Runtime} \\
\cmidrule(lr){2-3}\cmidrule(lr){4-5}\cmidrule(lr){6-7}\cmidrule(lr){8-8}
& $S_{\text{loc}}^*$ (\%) $\uparrow$ & $E_{\text{loc}}$ $\downarrow$
& $S_{\text{gra}}$ (\%) $\uparrow$ & $E_{\text{gra}}$ $\downarrow$
& $Q_{\text{smooth}}$ $\uparrow$ & $Q_{\text{line}}$ $\uparrow$
& $R_{\text{time}}$ $\uparrow$ \\
\midrule

\rowcolor{gray!08}
GT & 100.00 & 0.16 & 100.00 & 0.09 & 0.90 & 0.96 & 0.75 \\
\midrule

\multicolumn{8}{l}{\textbf{Closed-source VLMs}} \\
Gemini-3~Pro       & \textbf{7.00} &  \underline{4.09} & \textbf{2.00} & \textbf{3.98} & \underline{0.56} & 0.76 & \textbf{0.02} \\
GPT-5.1               &  \underline{4.00} & \textbf{4.02} & \textbf{2.00} & \underline{4.05} & 0.49 & \underline{0.81} & \textbf{0.02} \\
Qwen3-Max    & 3.00 & 4.70 & \underline{1.00} & 4.43 & \textbf{0.69} & \textbf{0.96} & \underline{0.01} \\
Grok4.1    & 0.00 & 5.33 & 0.00 & 5.29 & 0.51 & 0.68 & 0.00 \\
\midrule

\multicolumn{8}{l}{\textbf{Open-source VLMs}} \\
Qwen3-VL             & 3.00 & 4.91 & \underline{1.00} & 4.41 & 0.00 & 0.00 & \underline{0.01} \\
InternVL-3.5          & 0.00 & 5.06 & 0.00 & 5.10 & 0.46 & 0.50 & \underline{0.01} \\

\bottomrule

\end{tabular}
}
\vspace{-6pt}
\label{tab:exp:vlm}
\end{table}



As shown in~\Cref{tab:exp:vlm}, \emph{motion-aware localization is the core bottleneck}: all VLMs obtain only $S_{loc}\le 7\%$ with large errors ($E_{loc}\approx 4$--$5$), indicating persistent perception--timing mismatch when intercepting moving targets. Even conditioned on successful localization, grasping remains rare ($S_{gra}\le 2\%$, $E_{gra}\approx 4$--$5$), suggesting that contact-rich closure/stabilization is another major failure mode rather than a trivial “last step”. Model behaviors further decouple \emph{path quality} from \emph{task success}: Qwen3-Max achieves the best trajectory metrics ($Q_{smooth}=0.69$, $Q_{line}=0.96$) yet still has low $S_{loc}/S_{gra}$, while Gemini-3 Pro leads success ($S_{loc}=7\%$, $S_{gra}=2\%$) but with similarly high endpoint errors. 
\paragraph{Effectiveness of ObAct Baseline.}
\label{sec:baseline}
\begin{figure*}[t]
\centering
    \includegraphics[width=\linewidth]{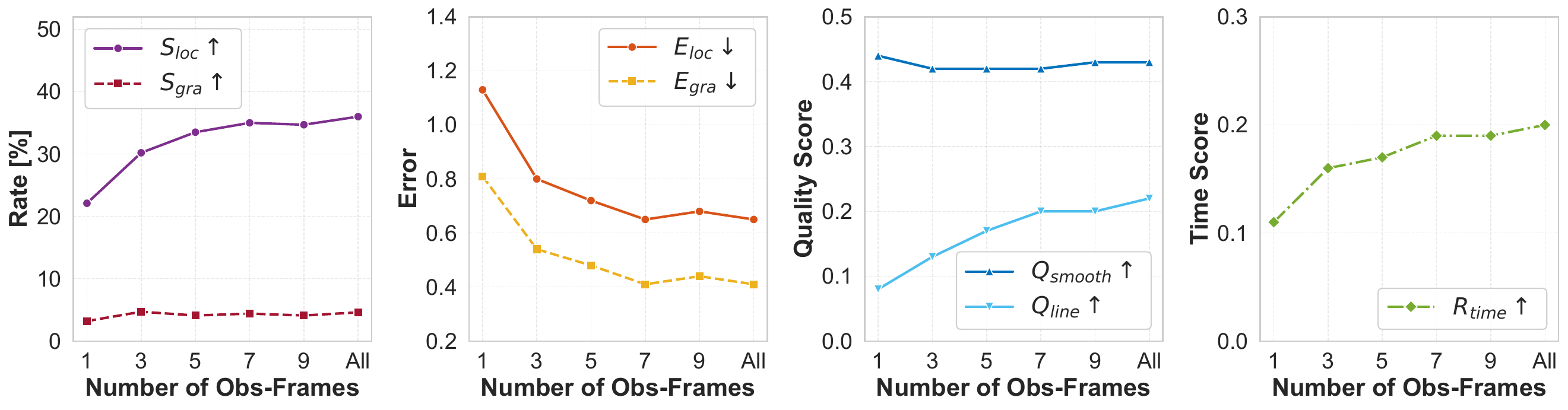}
    \caption{
    \textbf{Obs-frames ablation}. Scaling the number of sampled frames in the observation window steadily improves localization/grasping and reduces endpoint errors, with minor changes in trajectory quality.
    }
    \label{fig:baseline}
\end{figure*}

\Cref{fig:baseline} further studies the effect of temporal context by varying the number of observation frames. More observation frames yield consistent improvements (higher success rate and lower error), confirming that additional temporal evidence directly translates into better anticipation and capture timing. 
In contrast, trajectory metrics ($Q_{\text{smooth}}$, $Q_{\text{line}}$) change only modestly, suggesting they are intrinsic to the policy architecture (diffusion vs. autoregressive) rather than temporal context size.


\subsection{Detailed Analysis}
\label{sec:analysis}

\paragraph{Do Policy Models Learn to ``Grasp"?}
\begin{figure}[t]
\centering
    \includegraphics[width=1\linewidth]{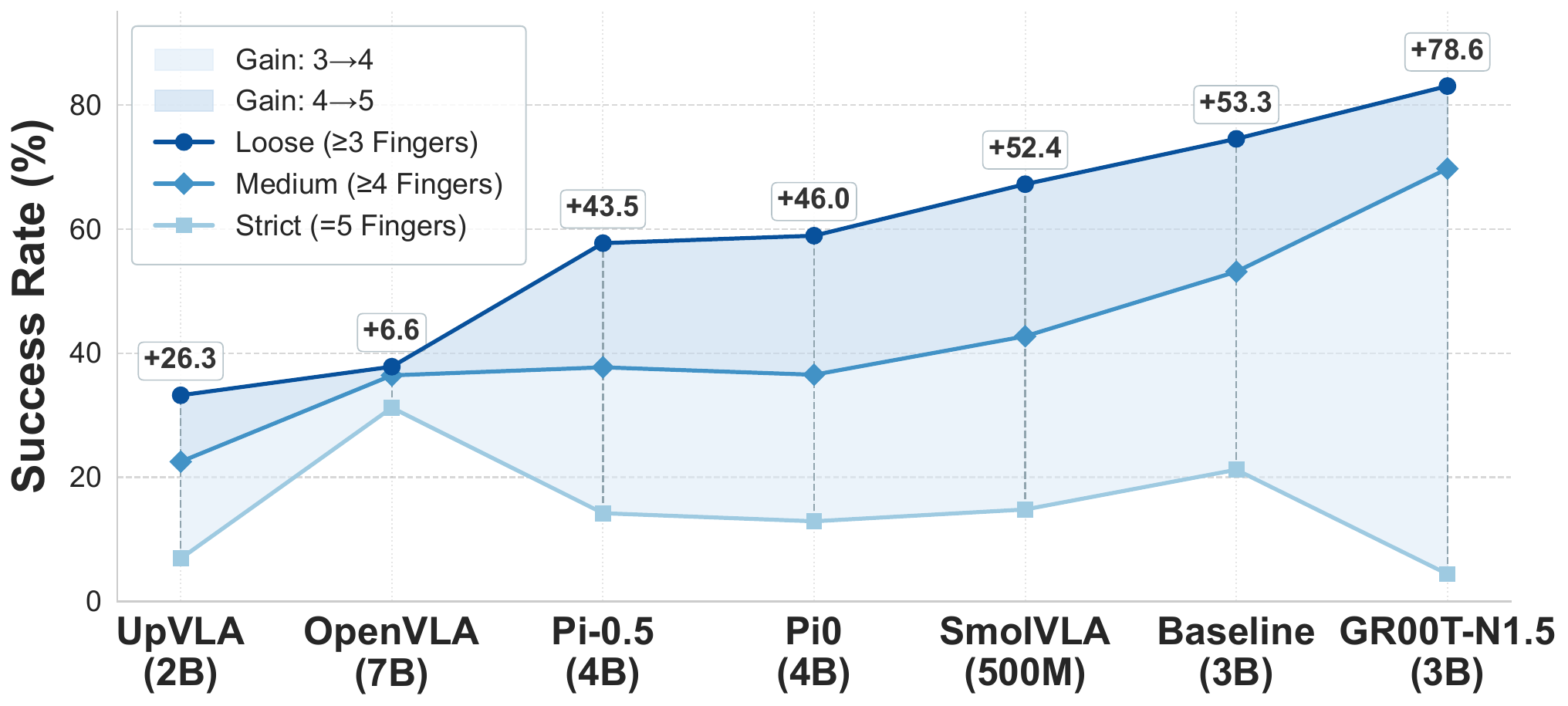}
    \caption{
    \textbf{Grasp success rate under per-frame evaluation.} Success rate over the trajectory under Loose/Medium/Strict settings (requiring at least 3, 4 and 5 fingers holding target).
    }
    \label{fig:losse_grasping}
    \vspace{-6pt}
\end{figure}

As discussed in \Cref{sec:evaluation_metrics}, grasping success is assessed only at the \emph{first localization-success} frame, i.e., the hand pose is judged for capture exactly when localization is achieved, which makes the metric stringent and often yields low $S_{\text{gra}}$ scores. 
We therefore perform an additional analysis over the full rollout by checking, at \emph{every} frame, whether the instantaneous grasp action is sufficient to capture the target. 
As illustrated in \Cref{fig:losse_grasping}, we define three difficulty levels and observe consistently higher per-frame success rate under relaxed criteria. Notably, diffusion policies yield significant gains when relaxing finger requirements (5$\to$4$\to$3). This suggests their grasps are locally consistent but lack the global synchronization required for strict, multi-finger success.

Even under the Loose criterion, most models struggle to surpass a 60\% success rate, whereas GR00T-N1.5 reaches $78.6\%$. This temporal inconsistency helps explain the low $S_{\text{gra}}$ in \Cref{tab:exp:vla}: grasp imitation is already inconsistent over time, making it unlikely to achieve a correct grasp precisely at the localized interception frame.

Finally, grasp-action consistency exhibits weak correlation with model size, echoing recent findings that VLA performance scales non-monotonically with parameter count~\cite{liurobomamba,wen2024tinyvla,wang2025vlaadapter}.

\paragraph{Data Scaling.}

\begin{figure}[t]
\centering
    \includegraphics[width=\linewidth]{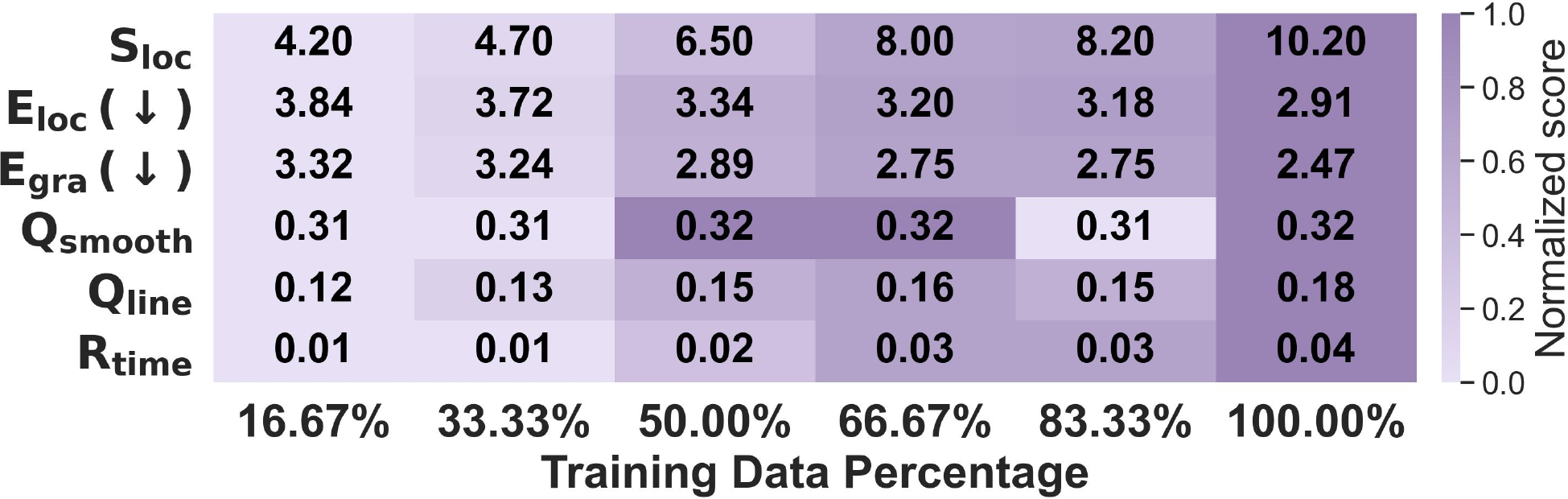}
    \caption{
    \textbf{Training-data scaling.} Test metrics vs.\ training data percentage (higher is better after normalization).
    }
    \label{fig:table_heatmap}
    \vspace{-6pt}
\end{figure}

As shown in \Cref{fig:table_heatmap}, task-level performance scales consistently with $S_{\text{loc}}$ rises steadily, while $E_{\text{loc}}$ and $E_{\text{gra}}$ decrease near-monotonically, indicating progressively more accurate motion-aware interception and capture execution. The smooth scaling trends across all task metrics further suggest that \benchname~provides high-quality and well-aligned supervision; substantial label noise or distributional shifts would typically manifest as unstable or non-monotonic behavior rather than sustained gains.

In contrast, trajectory-quality metrics exhibit early saturation: $Q_{\text{smooth}}$ and $Q_{\text{line}}$ vary only mildly across data scales, implying that additional data primarily benefits \emph{when and where} to intercept and grasp, rather than further refining path smoothness or spatial linearity. These geometry-level properties are instead more strongly governed by model inductive biases and architectural choices (e.g., diffusion-based vs.\ auto-regressive).

\section{Conclusion}

We presented the first evaluation platform and benchmark dedicated to Dynamic Capture: the \textbf{\suitename}~and the large-scale \textbf{\benchname}~benchmark. Our framework addresses the unique challenges of dynamic scenarios, moving targets, temporal alignment, and motion prediction, while supporting unified evaluation of VLAs, diffusion policies, and VLMs. We further establish an ObAct baseline that integrates short-horizon temporal observation with spatiotemporal attention, and show it consistently outperforms prior diffusion-based policies across all seven metrics (e.g., $+8.1\%$ $S_{\text{loc}}$). Our results show that despite strong performance in static settings, current models still struggle to anticipate and act under dynamic conditions. \suitename~and \benchname~shift the focus to dynamic grasping, advancing research toward motion-aware hand intelligence.

\section*{Impact Statement}




Dynamic hand–object interaction is relevant to many embodied systems where agents must act under changing conditions. This work contributes an evaluation-focused benchmark and protocol that can support more consistent comparisons across methods and help diagnose when models succeed or fail. Such evaluation infrastructure may benefit research on dexterous manipulation and perception–action integration, and can inform downstream applications (e.g., assistive robotics) once combined with appropriate engineering and validation. At the same time, improvements in physical interaction capabilities can be misused in harmful or unauthorized settings. We encourage responsible use, including safety constraints in development, careful real-world testing, and domain-specific safeguards before deployment.

\bibliography{example_paper}
\bibliographystyle{icml2026}

\appendix
\onecolumn

\section{Kinematic Primitive Foundations}~\label{app:theory}
We model moving targets in \benchname~using a compact vocabulary of basic kinematic primitives—linear (\texttt{L}), circular-arc (\texttt{A}), and simple-harmonic (\texttt{H}) motion. These three families strike a practical balance between expressiveness and controllability: they admit clean physical parameters (e.g., velocity/acceleration, curvature/radius, frequency/phase), support systematic difficulty scaling, and serve as composable “motion tokens” from which richer dynamics can be built. The remainder of this appendix formalizes this choice by stating our introducing notation for trajectories and curvature, and providing complementary signal-theoretic and geometric intuitions.


\paragraph{Notation.}
A trajectory is a differentiable curve $\mathbf r:[0,T]\!\to\!\mathbb R^d$ with velocity $\mathbf v(t)=\dot{\mathbf r}(t)$ and acceleration $\mathbf a(t)=\ddot{\mathbf r}(t)$. For planar curves parameterized by time, curvature is
\begin{equation}
\kappa(t)=\frac{\|\mathbf v(t)\times \mathbf a(t)\|}{\|\mathbf v(t)\|^3},
\end{equation}
and in the arc-length parameter $s$ we use the Frenet-Serret frame $(\mathbf T,\mathbf N,\mathbf B)$.
For plane curves, the \emph{fundamental theorem of curves} states that the curvature function $\kappa(s)$ determines the curve up to a rigid motion. We use $\tau(s)$ for torsion in 3D when needed. 

\paragraph{Signal-theoretic (harmonic) view.}
Treat each coordinate of $\mathbf r(t)$ as a scalar time signal. If $x(t)\in L^2([0,T])$ is $T$-periodic, then by the Riesz-Fischer theorem its Fourier series converges in the $L^2$ sense:
\begin{equation}
x(t)=a_0+\sum_{k=1}^{\infty}\big(a_k\cos(k\omega t)+b_k\sin(k\omega t)\big),\ \ \omega=\tfrac{2\pi}{T}.
\end{equation}
Hence $\mathbf r(t)$ can be approximated arbitrarily well (in energy) by a finite trigonometric polynomial.
Uniform-frequency sine/cosine pairs with phase shift realize circular motions in $(x,y)$, the affine (zero-frequency) part gives linear drift.
This motivates a small set of \emph{frequency tokens} (dominant $\{k, A_k,\phi_k\}$ over local windows) to summarize oscillatory micro-motions under a controllable truncation budget.

\paragraph{Geometric (curvature) view.}
Because planar shape is encoded by $\kappa(s)$, approximating $\kappa(s)$ by a step function yields a piecewise constant-curvature curve, \ie, a concatenation of straight segments ($\kappa\!\approx\!0$) and circular arcs ($\kappa\!\approx\!1/R$).
This discretization aligns with nonholonomic optimal-control results:
(i) \textbf{Dubins}: with bounded curvature and forward-only motion, the shortest path between two poses consists of at most three pieces from $\{L,R,S\}$ (left/right arcs and straight lines). 
(ii) \textbf{Reeds--Shepp}: allowing reversals, shortest paths admit finite canonical sequences of arcs/segments. 
When steering-rate continuity is required, \emph{clothoids} (Euler spirals, linearly varying curvature) provide $G^2$-friendly connectors between $L/A$ pieces. 
These facts justify a compact \emph{geometry-token} vocabulary with interpretable parameters (length, radius/sign, curvature slope). \citep{dubins1957,reeds1990,lavalle2006}

\section{Motivation Claim}~\label{app:motivation}

Current work on hand action generation mainly focuses on manipulating static objects, with limited models and benchmarks for dynamic object interaction. In static grasping tasks, evaluation can rely directly on ground-truth (GT) trajectories since the target remains fixed and a unique reference exists. In dynamic tasks, however, the agent may choose to grasp the moving object at different valid time points, and even if a GT is defined, it represents only one of many feasible solutions. This necessitates an interactive environment to assess whether a model can capture motion patterns and successfully grasp within the effective time window. To this end, we introduce the \textbf{\suitename}, a unified platform for data generation and interactive evaluation in dynamic hand grasping.

\section{Limitations and Future Work}
\label{app:limitation}
\paragraph{Limitations.}
Despite the strengths of DynaHOI-Gym and DynaHOI-10M in evaluating dynamic capture under controlled target motion, the current study has several limitations:

\begin{enumerate}
    \item \textbf{Limited appearance and material diversity.}
    The benchmark prioritizes motion diversity and temporal alignment, but the range of object appearances and material properties remains limited. While multiple shapes and scales are included, challenging visual factors such as transparency, deformability, and strong specular reflections are not yet systematically modeled.

    \item \textbf{Simplified perception setting.}
    Observations are restricted to single-view RGB inputs with fixed camera poses. This design isolates motion-aware reasoning but simplifies perception compared to real-world scenarios involving multi-view sensing, viewpoint changes, and severe hand–object occlusions.

    \item \textbf{Entangled localization and grasping failures.}
    The evaluation focuses on end-to-end action generation, revealing that successful localization does not reliably translate into successful grasping. However, the current setup does not explicitly disentangle the failure modes of localization and contact-rich grasp execution.
\end{enumerate}

\paragraph{Future Work.}
These limitations suggest several promising directions for future research:

\begin{enumerate}
    \item \textbf{Stage-wise coupling of localization and grasping.}
    Our results indicate that grasping frequently fails even when localization succeeds at the interception frame, pointing to a weak coupling between spatial alignment and finger closure.
    A natural extension is to decompose dynamic capture into explicit stages, where successful localization triggers a dedicated grasping phase.
    For example, once the palm enters a valid interception region, the policy could receive an explicit transition signal or conditioning cue that shifts its objective from motion anticipation to contact stabilization, thereby more tightly linking localization and grasp execution.

    \item \textbf{Structured multi-stage pipelines beyond end-to-end policies.}
    While this work focuses on evaluating zero-shot, end-to-end models, future studies may explore more structured multi-stage approaches.
    Such pipelines could integrate specialized components, including explicit depth or geometry estimation for spatial reasoning, short-horizon motion prediction for target forecasting, and dedicated hand controllers that output coordinated wrist and finger trajectories.
    Although these designs may reduce generality, they provide stronger inductive biases for precise interception and contact-rich manipulation.

    \item \textbf{Benchmarking modular and hybrid designs.}
    DynaHOI-Gym offers a controlled and reproducible testbed for systematically comparing end-to-end policies with stage-wise or hybrid systems.
    Future work can leverage this platform to analyze how modular structure affects robustness, timing accuracy, and grasp reliability under dynamic conditions.
\end{enumerate}

\section{Data Collection and Processing.}~\label{app:data_collection}

\paragraph{Data collection.}
During data collection, we run the Unity simulation with fixed-step logging at 20\,Hz (i.e., \texttt{sendInterval}$=0.05$s). For each episode, an egocentric hand-mounted camera renders to a dedicated \texttt{RenderTexture} of size \texttt{width}$\times$\texttt{height}, and every logged frame is saved as a JPEG image (\texttt{EncodeToJPG(85)}) to \texttt{episode\_\{id\}/img\_\{t:04d\}.jpg}. Synchronously, we record the full hand kinematic state as a per-frame JSON file \texttt{episode\_\{id\}/joints\_\{t:04d\}.json}. Specifically, we serialize a set of tracked transforms \texttt{storeList} that includes (i) the 15 actuated finger joints, (ii) five fingertip ``nail'' keypoints, (iii) the wrist root (\texttt{handsRoot}), (iv) the palm-center marker (\texttt{palmCenter}), and (v) the egocentric camera itself. For each tracked transform, we store its 3D world position and 3D orientation in Euler angles (\texttt{joint.eulerAngles}), yielding a time-aligned sequence of RGB observations and hand poses.
Episodes are parameterized by motion scripts, whose per-episode configurations are loaded from JSON via an \texttt{EpisodeManager}. At the beginning of each episode (saved once), we also write \texttt{meta\_data.json} containing the task type, motion parameters for the active script, the predicted intercept position $\texttt{targetPosition}$ (computed from the motion model using $\texttt{interceptTime}$ with an anticipation margin $\texttt{leadTime}$ and an observation delay $\texttt{observationTime}$), and the resulting hand translation speed $\texttt{moveSpeed}$ required to reach the intercept point in time.
The hand executes an observe-before-act state machine: after an observation wait, it moves toward the predicted intercept, optionally waits for the target to enter a small planar tolerance, and then closes the fingers until either all five fingers make contact (detected by sphere-overlap tests on collision-enabled joints) or a maximum joint-rotation limit is reached ($\texttt{maxGrabRotation}=90^\circ$). To improve data integrity, failed simulations are automatically retried up to three times; on each retry we clear previously saved frames for that episode and re-run the same configuration, while successful episodes are explicitly finalized and advanced to the next configuration.

\paragraph{Post-processing and quality control.}
After collection, we convert the recorded pose streams into action sequences at 20\,Hz (e.g., joint-rotation increments for the 15 actuated joints, optionally augmented with palm/wrist motion depending on the downstream policy interface), and compute dataset-wide descriptive statistics for each action dimension, including \texttt{min}, \texttt{max}, and robust quantiles (\texttt{q01}, \texttt{q99}). We visualize per-dimension frequency histograms to inspect scale and saturation effects, then apply outlier filtering by clipping or removing samples outside the robust range (e.g., beyond $[\texttt{q01},\texttt{q99}]$ with a small safety margin) and by rejecting entire trajectories that exhibit systematic corruption (e.g., persistent saturation at limits, discontinuous jumps inconsistent with the control rate, or missing/invalid frames). Finally, we perform stratified random sampling for manual inspection: we replay sampled episodes by synchronizing RGB frames with the recorded kinematics and actions to verify temporal alignment, motion plausibility, and successful grasp--release behavior, and we remove any remaining ``dirty'' trajectories that fail these checks.

\section{Testing Details on \suitename}
\label{app:TestDetail}

\subsection{\suitename~Settings}
\begin{figure*}[t]
    \centering
    \setlength{\tabcolsep}{2pt}

    \begin{tabular}{ccccc}
        \subcaptionbox{Centered\label{fig:offset_center}}
        {\includegraphics[width=0.15\textwidth]{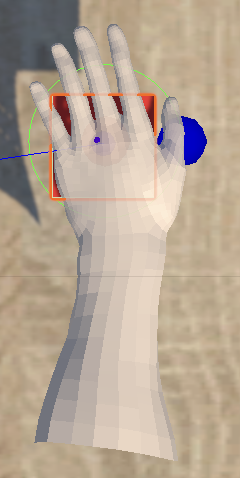}} &
        \subcaptionbox{$Z$: +0.3\label{fig:offset_z_pos}}
        {\includegraphics[width=0.15\textwidth]{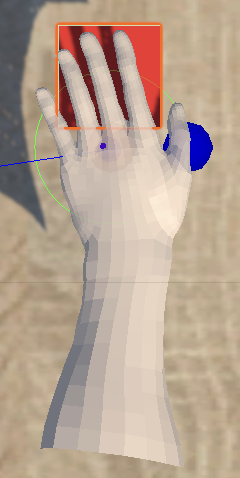}} &
        \subcaptionbox{$Z$: --0.3\label{fig:offset_z_neg}}
        {\includegraphics[width=0.15\textwidth]{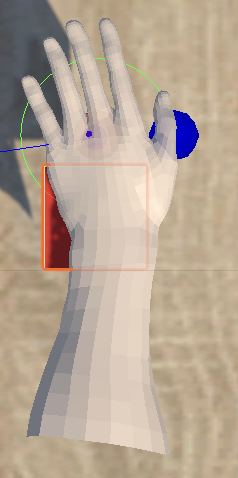}} &
        \subcaptionbox{$X$: +0.3\label{fig:offset_x_pos}}
        {\includegraphics[width=0.15\textwidth]{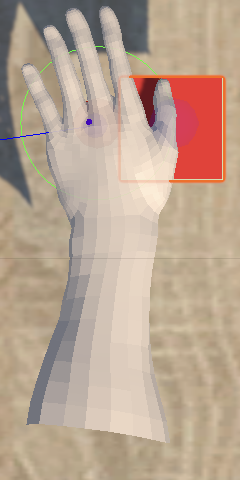}} &
        \subcaptionbox{$X$: --0.3\label{fig:offset_x_neg}}
        {\includegraphics[width=0.15\textwidth]{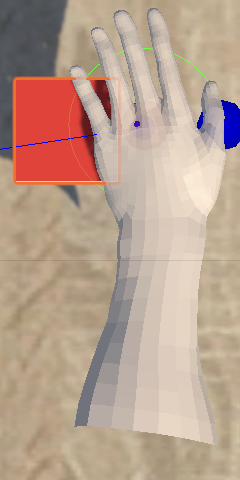}}
    \end{tabular}

    \caption{\textbf{Qualitative visualization of target offsets.}
    We render five representative views under different offset settings (centered, $\pm 0.3$ along the $Z$ axis, and $\pm 0.3$ along the $X$ axis), illustrating how the target region shifts relative to the hand.}
    \label{fig:offset_qual}
\end{figure*}

During evaluation, the policy model communicates with \suitename~in real time. At each time step, the simulator provides the latest observation (image and hand state), which is immediately forwarded to the model for inference, and the predicted action is then transmitted back to the simulator for execution.

Although we configure the simulator to maintain strict synchronization between the model and the simulated environment (ensuring that object and hand positions in the captured images are accurate), the real-time rendering process introduces subtle pixel-level variations across repeated trials at the same timestep. These variations are imperceptible to the human eye yet can slightly perturb the model’s outputs. This setting imposes a stricter robustness requirement: while the critical information (object location and hand position) remains consistent, the model should not produce unstable or divergent actions due to visually negligible background differences.

\subsection{Evaluation Criteria}
\begin{enumerate}[leftmargin=*]
    \item Localization success: we deem the palm center successfully localized if its distance to the object center is below 0.3 Unity world units, a threshold that ensures the palm can cover at least half of the object area (see \Cref{fig:offset_qual}). For VLM evaluation, we relax this threshold to 1.0 Unity units to avoid near-zero success rates and improve score sensitivity.
    \item Grasping success: Following the grasping abstractions in Habitat 2.0 and the attachment mechanisms in MuJoCo (weld constraints) and Isaac Sim (surface grippers), we automatically attach the target object to the hand once the hand–object distance meets the localization criterion. A grasp is then deemed successful if the current grasp configuration matches the GT sufficiently well: for each of the 15 joints, the predicted rotation has the correct direction and reaches at least $0.9\times$ the GT rotation magnitude.
\end{enumerate}

While Unity’s real-time rendering engine ensures high physical realism, its inherent nondeterminism poses challenges to evaluation consistency. To address this, we employ Unity’s internal clock for environment stepping during data collection, ensuring realistic yet diverse sampling. During model testing, the platform advances the simulator manually in a timed manner, synchronizing environment updates with model inference on the temporal axis. This mechanism guarantees time consistency between the environment and the model, enabling \suitename~to serve as a reliable online evaluation platform and providing a unified benchmark for dynamic hand grasping research.

\subsection{Details for Evaluating VLMs}
\paragraph{Online evaluation protocol.}
We evaluate VLMs in a closed-loop Unity simulator via a WebSocket interface.
For each episode, the Python controller first reads the episode metadata (episode id, task type, and episode length $L$) from a LeRobot-formatted dataset.
It then sends a \texttt{start\_episode} signal to Unity, including the action horizon $T$ (we use $T{=}10$).
During the rollout, Unity repeatedly streams the current observation as \texttt{image\_and\_state}, consisting of an RGB image and the current 18D hand state $\mathbf{s}_t\in\mathbb{R}^{18}$ (3D hand position plus 15 finger-joint angles).
At each step, the VLM receives \emph{(image, prompt)} and predicts the next $T$ frames of 18D hand parameters, which are sent back to Unity as \texttt{action\_data}; if the remaining episode length is shorter than $T$, we truncate the action chunk to match the episode boundary.
The episode terminates when Unity returns a \texttt{metrics} message, from which we record success and distance-based scores, and we additionally log the raw VLM outputs for auditing.

\paragraph{Prompt format.}
We use a fixed structured prompt that specifies (i) the task instruction, (ii) the current 18D hand state, (iii) the prediction horizon $T$, and (iv) a strict JSON-only output requirement to ensure reliable parsing.

\begin{figure}[t]
    \centering
    \begin{mdframed}[
        linewidth=0.6pt,
        innerleftmargin=6pt,
        innerrightmargin=6pt,
        innertopmargin=6pt,
        innerbottommargin=6pt
    ]
    \small
\textbf{System Role.} You are a Vision-Language-Action agent for dynamic hand--object interaction.\\
\textbf{Task.} \texttt{<TASK\_TEXT>}\\
\textbf{Input.} RGB image + current hand state $\mathbf{s}_t\in\mathbb{R}^{18}$:
\texttt{[x, y, z, 15 joint angles (rad)]}.\\
\textbf{Skill Set.} \texttt{\{WAIT, APPROACH, INTERCEPT, GRASP, LIFT, ADJUST\}}.\\
\textbf{Output.} Return (i) a parameterized skill program whose durations sum to $T$, and (ii) an explicit low-level rollout of the next $T$ frames, each with 18 values in the same order.\\[2pt]
\textbf{Return \emph{only} the following JSON:}

\begin{Verbatim}[fontsize=\small]
{
  "action_sequence": [
    {
      "skill": "APPROACH",
      "params": {"target": "object",......},
      "duration": 4,
      "terminate_if": ...,
    },
    {
      "skill": "GRASP",
      "params": {...},
      "duration": 6
    }
  ],
  "predicted_motion": [
    {"frame_index": 1, "hand_params": [18 values]},
    ...,
    {"frame_index": T, "hand_params": [18 values]}
  ]
}
\end{Verbatim}
    \end{mdframed}
    \caption{Hybrid prompt template for VLM online control. \texttt{<TASK\_TEXT>} specifies the motion instruction, and $T$ is the action horizon (default $10$). The model outputs a parameterized skill program (\texttt{action\_sequence}) and a $T$-step low-level control rollout.}
    \label{fig:vlm_prompt_template}
\end{figure}

\section{Why $Q_{\text{line}}$ and $Q_{\text{smooth}}$ for GT are $<1.0$.}
~\label{app:Q_line_Q_smooth_explain}
In principle, the scripted hand trajectory during the approach phase is designed to be constant-speed translation, and thus an idealized ground-truth (GT) trajectory would yield $Q_{\text{line}}=Q_{\text{smooth}}=1.0$. In our Unity-based data generation, however, the recorded waypoints are sampled from the real-time simulation loop and therefore exhibit mild temporal jitter, which propagates into the geometry-based quality scores.

Concretely, joint states are logged inside \texttt{FixedUpdate()} using a time-gated condition of the form \texttt{Time.time - lastSendTime $\ge$ sendInterval}. This couples a physics-timestep callback (\texttt{FixedUpdate}) with the wall-clock-like game time (\texttt{Time.time}) rather than a strictly uniform sampling clock (e.g., \texttt{Time.fixedTime} with deterministic accumulation). As a result, the effective sampling intervals are not exactly constant: small fluctuations in frame scheduling lead to non-uniform timestamps, and thus the consecutive recorded positions may not be evenly spaced even when the underlying motion is close to constant-velocity.

This effect is further amplified by the per-sample I/O workload executed within \texttt{FixedUpdate()}, including camera rendering, GPU$\rightarrow$CPU readback (\texttt{ReadPixels}), JPEG encoding, and synchronous disk writes for images and JSON metadata. These operations introduce variable latency and occasional stalls, which perturb the cadence at which the logging condition is satisfied and yield uneven inter-sample spacing. Because $Q_{\text{line}}$ and $Q_{\text{smooth}}$ are computed from the discrete recorded trajectory (rather than from an analytic, continuous-time reference), such sampling irregularities manifest as small but systematic deviations from an ideal straight and smooth polyline, producing GT scores slightly below $1.0$ (e.g., $Q_{\text{line}}\approx0.96$ and $Q_{\text{smooth}}\approx0.90$). Importantly, this gap reflects instrumentation and scheduling artifacts of the simulation/recording pipeline rather than imperfections in the intended GT motion itself.

\section{Stratified Performance Analysis}
~\label{app:stratified}

\paragraph{Periodicity-stratified evaluation.}
We stratify test trajectories by \emph{periodicity strength}: (i) \textbf{Circular} exhibits the clearest periodic signature since each sampled frame provides phase-distinct visual evidence, making motion state more identifiable under sparse observations; (ii) \textbf{Periodic} (like harmonic and pendulum motions) are weaker-periodic, characterized by higher positional revisitation and larger low-speed occupancy (dwelling near turning points). It
reduces effective phase diversity given the same sampling budget; (iii) \textbf{Linear} (like straight line and projectile motions) are non-periodic.

\begin{figure}[t]
\centering
    \includegraphics[width=0.7\linewidth]{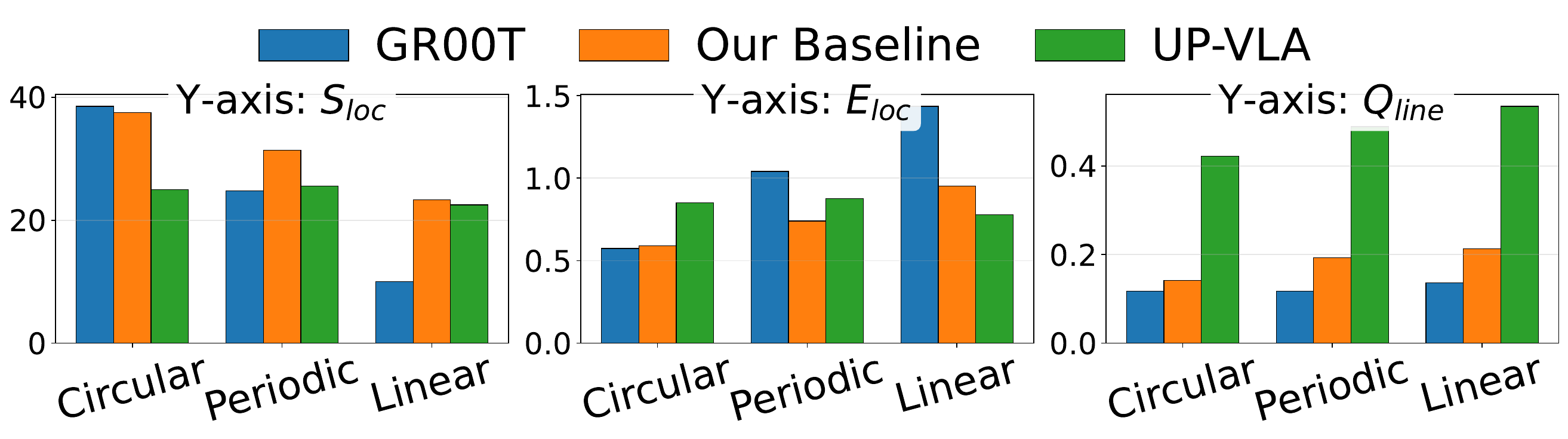}
    \caption{Periodicity-stratified performance.}
    \label{fig:trend_by_type}
\end{figure}

As shown in~\Cref{fig:trend_by_type}, diffusion-based policies show a pronounced periodicity sensitivity: they achieve higher $S_{\text{loc}}$ and lower $E_{\text{loc}}$ on more periodic motions, suggesting stronger reliance on recognizable temporal patterns for interception timing. For trajectory quality, both AR-based and diffusion-based models improve on more periodic motions, yielding higher $Q_{\text{line}}$, indicating more stable, motion-consistent control when dynamics are more predictable. 

Finally, $E_{\text{gra}}$ follows the same trend as $E_{\text{loc}}$, and $Q_{\text{smooth}}$ mirrors $Q_{\text{line}}$. Details are reported in Appendix~\ref{app:detail_periodicity}.

\begin{figure}[t]
\centering
    \includegraphics[width=0.7\linewidth]{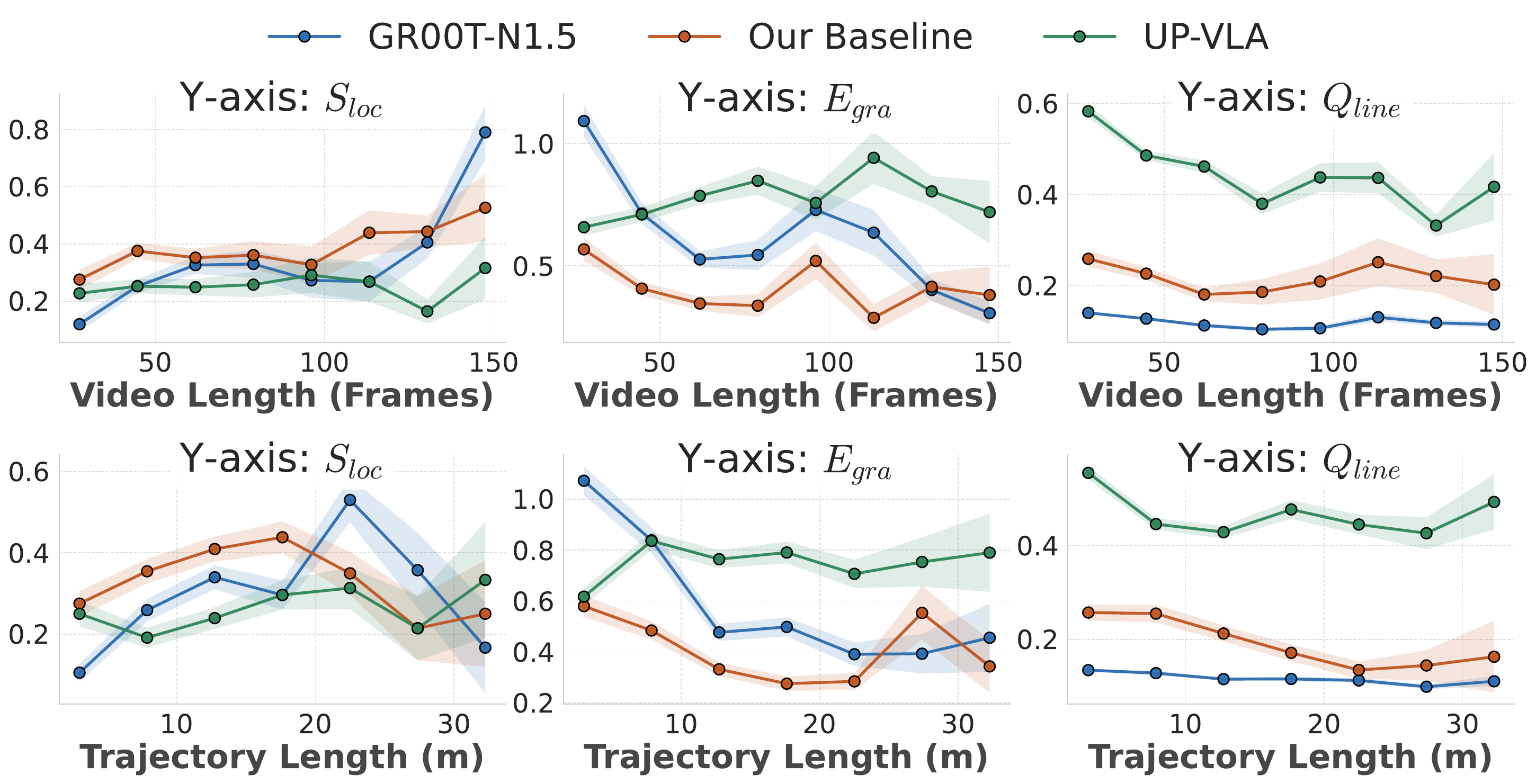}
    \caption{
    Length-stratified performance.
    }
    \label{fig:trend_by_length}
\end{figure}

\paragraph{Length-stratified trends.}
As shown in \Cref{fig:trend_by_length}, diffusion-based policies exhibit the clearest length sensitivity on \emph{task outcomes}: $S_{\text{loc}}$ increases with longer observation (video length), while $S_{\text{gra}}$ decreases accordingly; notably, the gain in $S_{\text{loc}}$ becomes non-monotonic when conditioned on trajectory length. In contrast, grasping endpoint error shows a more consistent improvement: $E_{\text{gra}}$ decreases markedly as both video length and trajectory length grow, indicating more accurate contact execution under longer temporal context and longer-horizon motion. 

For both diffusion- and AR-based models, trajectory quality is largely length-invariant: $Q_{\text{smooth}}$ and $Q_{\text{line}}$ show no systematic correlation with either video length or trajectory length, suggesting that path regularity is governed more by the policy/decoder bias than by temporal/trajectory scale. Finally, the paired metrics co-vary as expected: $S_{\text{gra}}$ tracks $S_{\text{loc}}$, $E_{\text{gra}}$ mirrors $E_{\text{loc}}$, and $Q_{\text{smooth}}$ mirrors $Q_{\text{line}}$; full trend plots are provided in Appendix~\ref{app:length_stratified}.

\section{Details about Periodicity-Stratified Evaluation}
\label{app:detail_periodicity}

\begin{figure*}[h]
\centering
    \includegraphics[width=0.9\linewidth]{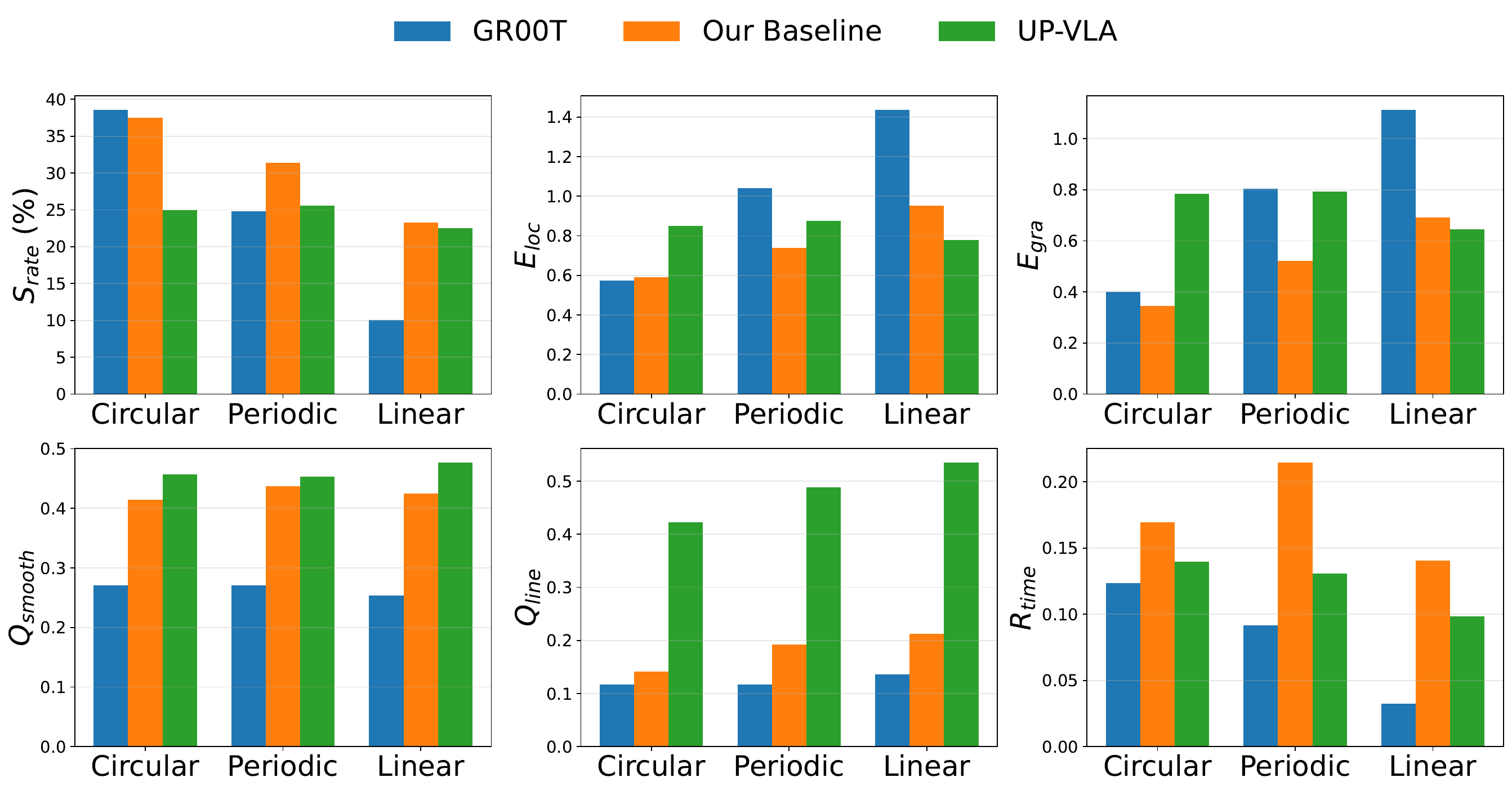}
    \caption{
    All periodicity-stratified performance.
    }
    \label{fig:trend_by_type_all}
\end{figure*}

\newpage
\section{Details about Length-Stratified Trends}
\label{app:length_stratified}

\begin{figure}[!h]
\centering
    \includegraphics[width=\linewidth]{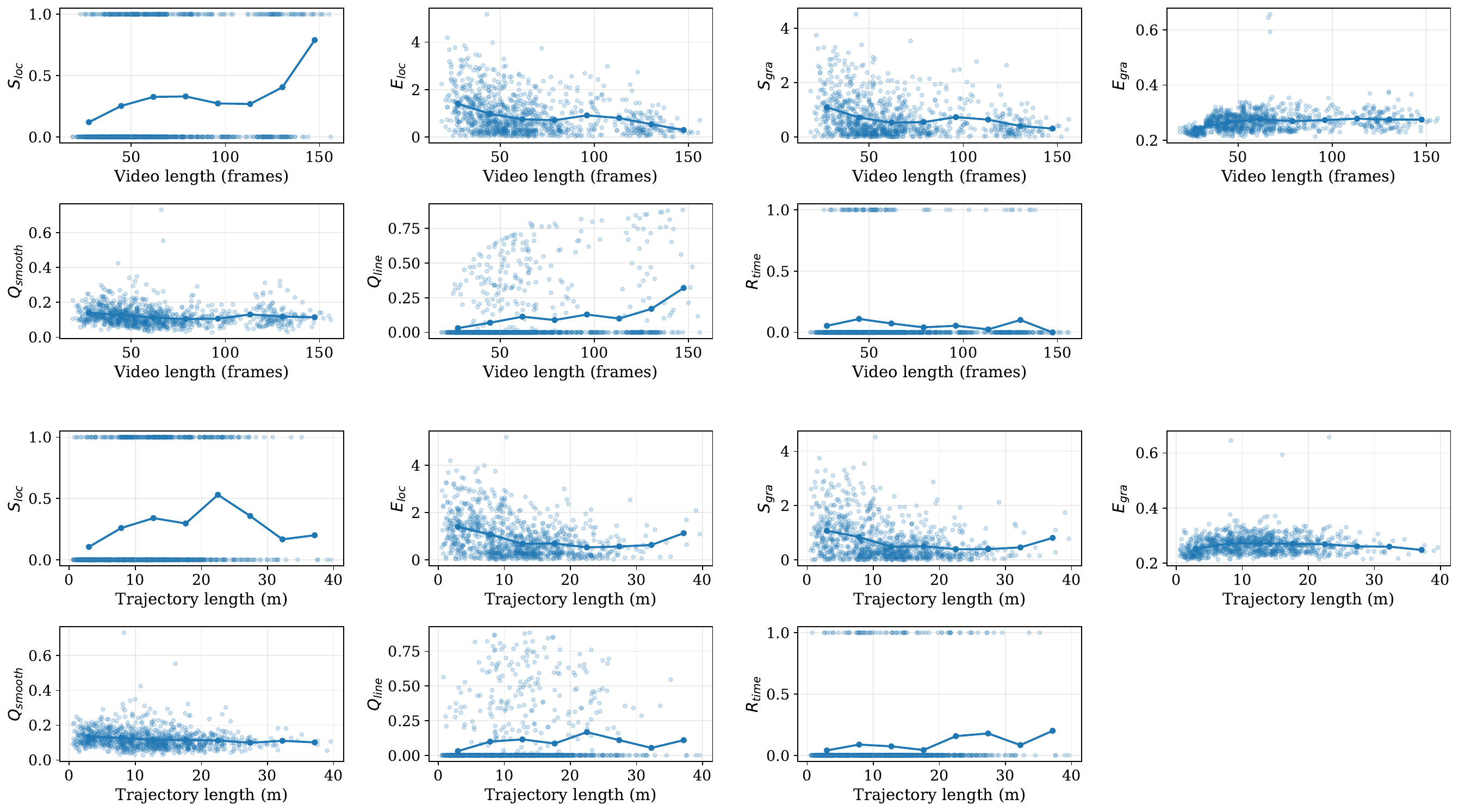}
    \caption{
    All length-stratified performance (GR00T-N1.5).
    }
    \label{fig:trend_by_length_all_gr00t}
\end{figure}

\begin{figure}[!h]
\centering
    \includegraphics[width=\linewidth]{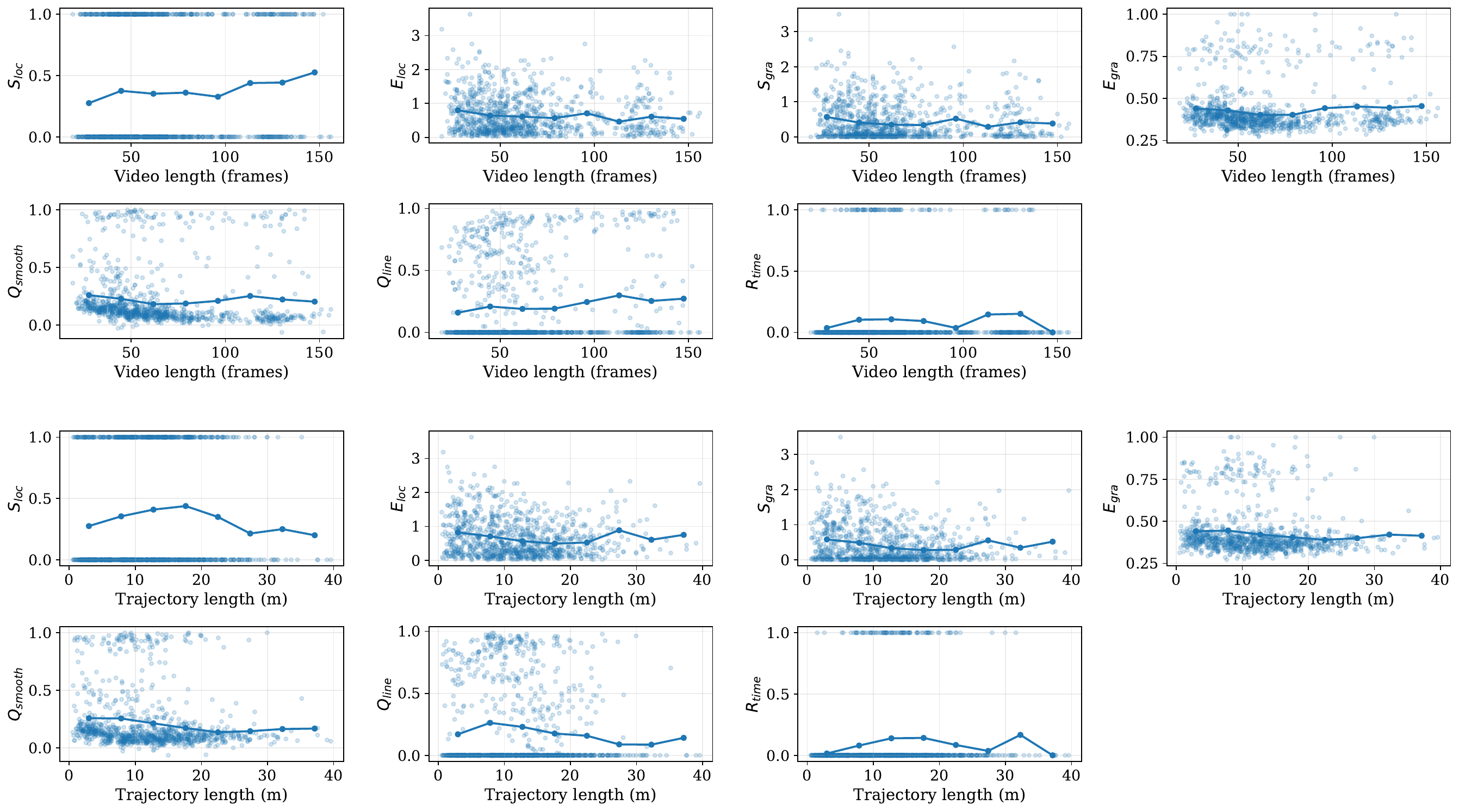}
    \caption{
    All length-stratified performance (our baseline).
    }
    \label{fig:trend_by_length_all_baseline}
\end{figure}

\newpage
\section{Comprehensive Distribution of Hand Motion}
\label{app:all_hand_motion}

\begin{figure}[!h]
\centering
    \includegraphics[width=\linewidth]{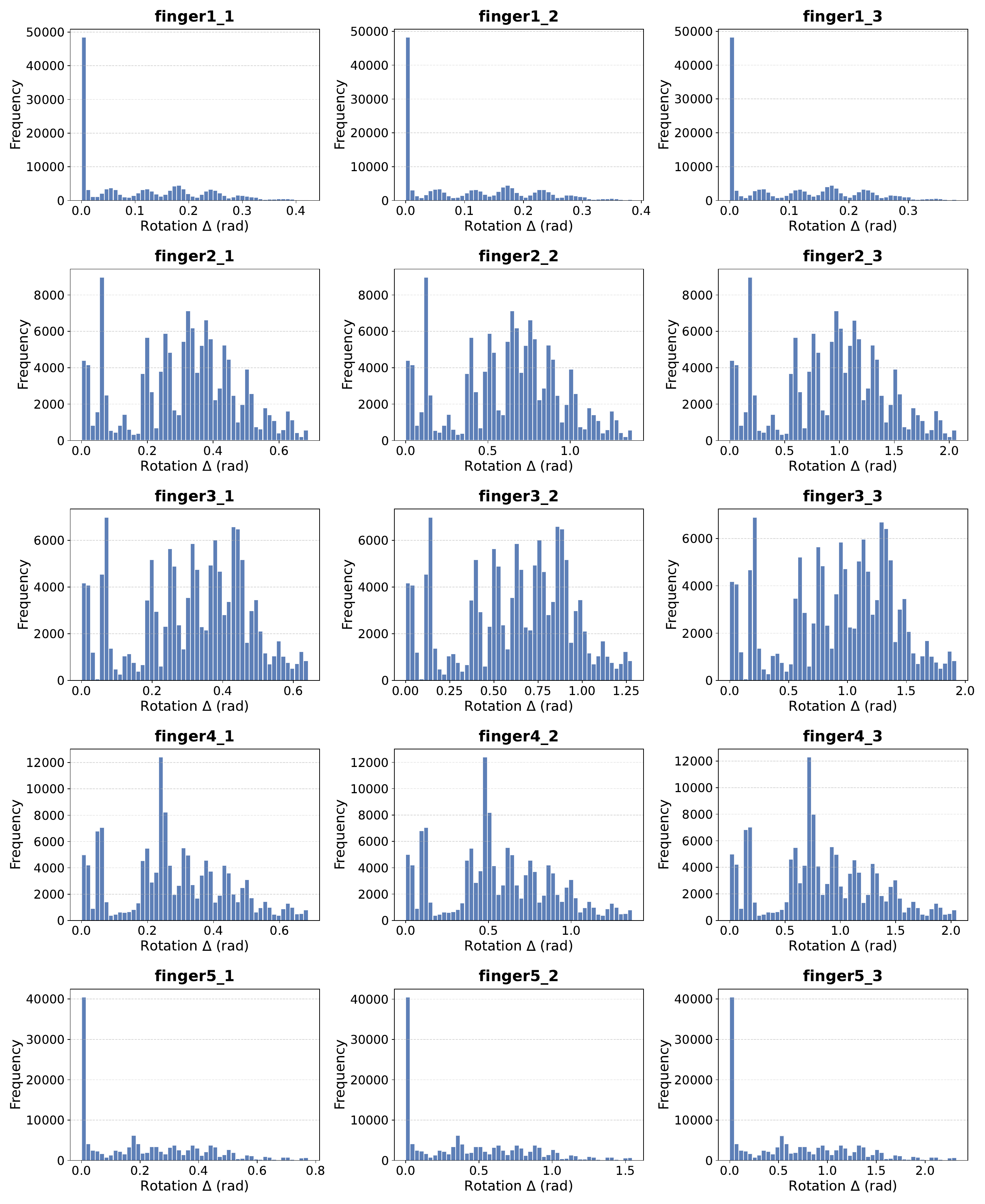}
    \caption{
    Distribution of 15-DoF joint rotation deltas. The histograms visualize the frame-to-frame rotation changes across all joints.
    }
    \label{fig:rotation_distribution}
\end{figure}

\begin{figure}[h]
\centering
    \includegraphics[width=\linewidth]{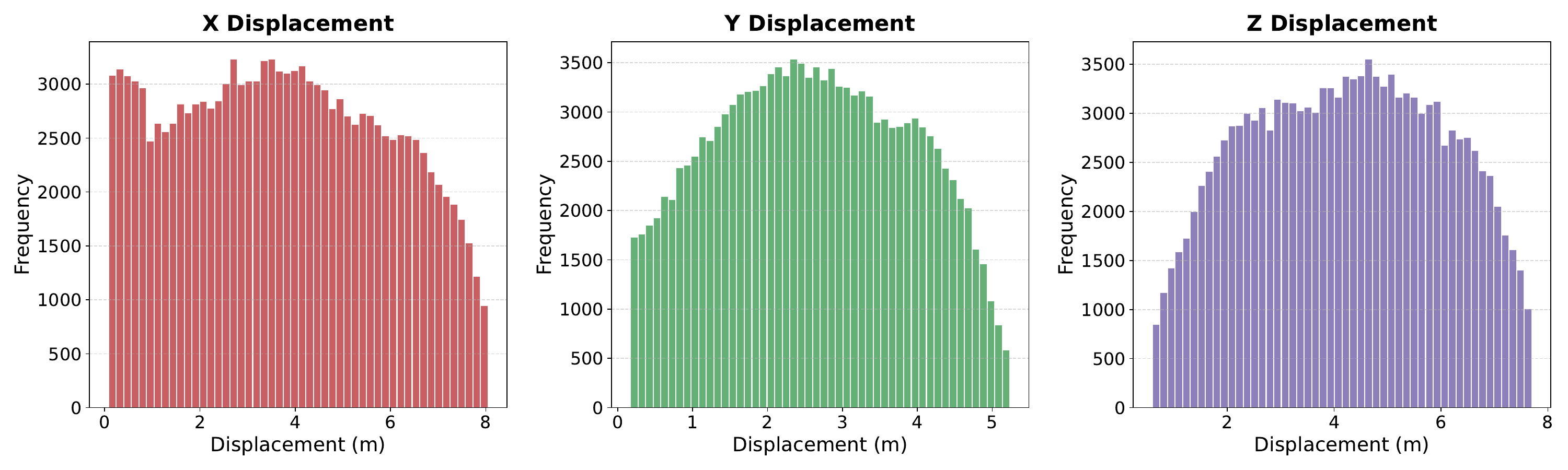}
    \caption{
    Distribution of 3D Cartesian displacements. The frequency of positional changes in X, Y, and Z directions for the hand root, demonstrating the spatial coverage of the benchmark.
    }
    \label{fig:displacement_distribution}
\end{figure}

\end{document}